\documentclass[preprint,10pt,times]{elsarticle}

\usepackage[english]{babel}
\usepackage{amssymb}
\usepackage{amsmath}
\usepackage{hyperref}
\usepackage{xcolor}
\usepackage{lineno}
\usepackage{microtype}
\usepackage{booktabs}
\usepackage{subcaption}
\usepackage{placeins}
\usepackage{multirow}

\hypersetup{
    colorlinks=true
}

\journal{}

\begin{document}

\begin{frontmatter}



\title{Deep Reinforcement Learning for Heat Pump and Underfloor Heating Control: An End-to-End Case Study}


\author[UGR]{Antonio Manjavacas\corref{corr}}
\ead{manjavacas@ugr.es}

\author[GF]{David Suárez-Caro}
\ead{david.suarez@georgfischer.com}

\author[GF]{Alejandro Campoy-Nieves}
\ead{ext.alejandro.campoy@georgfischer.com}

\author[GF]{Guillermo Rey}
\ead{guillermo.rey@georgfischer.com}

\author[GF]{Gonzalo Satrústegui}
\ead{gonzalo.satrustegui@georgfischer.com}

\author[GF]{Diego Farto-Alonso}
\ead{diego.farto@georgfischer.com}

\author[UGR]{Miguel Molina-Solana}
\ead{miguelmolina@ugr.es}

\affiliation[UGR]{organization={Department of Computer Science and Artificial Intelligence},
            addressline={Universidad de Granada}, 
            city={Granada},
            postcode={18071}, 
            country={Spain}}

\affiliation[GF]{organization={Georg Fischer Building Flow Solutions},
            addressline={Avda. Leonardo da Vinci 15-17-19}, 
            city={Getafe},
            postcode={28906}, 
            state={Madrid},
            country={Spain}}

\cortext[corr]{Corresponding author.}


\begin{abstract}
Efficient control of underfloor heating (UFH) systems coupled with heat pumps (HPs) is challenging due to the high thermal inertia of buildings and the slow dynamic response of UFH systems. Deep Reinforcement Learning (DRL) offers a data-driven control approach capable of simultaneously optimizing energy consumption and thermal comfort. However, the application of DRL-based controllers in occupied buildings under real operating conditions remains limited. This work presents and end-to-end case study (development, evaluation, and real-world deployment) of a DRL-based controller for an integrated UFH–HP system. First, a simulation-based pilot study was conducted to assess different DRL algorithms, control strategies, and configurations. Subsequently, a digital twin of a real residential building was developed and used to train DRL agents under simulated operating conditions. The best-performing agent was then deployed in the physical building without additional online fine-tuning. The results show that the controller successfully maintained indoor temperatures within the prescribed comfort limits while dynamically adjusting the HP supply temperature and UFH valve positions. These findings demonstrate the feasibility of deploying DRL controllers in occupied buildings and highlight their robustness and the practical challenges associated with transferring control policies from digital environments to real heating systems.
\end{abstract}



\begin{keyword}
Reinforcement learning \sep HVAC \sep Underfloor heating \sep Radiant floor \sep Heat pump \sep Control


\end{keyword}

\end{frontmatter}



\section{Introduction}
\label{sec:intro}

Underfloor heating (UFH) systems provide a radiant heating source that ensures uniform thermal distribution, stable indoor temperatures, and enhanced thermal comfort. Among them, hydronic UFH systems coupled with air-to-water or water-to-water heat pumps (HPs) have emerged as a highly promising solution for efficient building climate control \cite{iea2023heat}. Moreover, their operation at lower supply water temperatures than conventional convective heating systems enables highly efficient integration with HPs, reducing thermal losses, improving overall energy performance, and lowering operating costs \cite{cho2023integrated,zelazna2025review,hwang2021energy}. In addition, the capability of HPs to integrate with renewable energy sources, smart grids, and energy storage systems further positions this combination as a key technology for the decarbonization of the building sector \cite{mugnini2025integrated,iea2023the,brudermueller2025estimation}.

Despite these advantages, UFH systems introduce operational challenges that are rarely encountered in conventional convective heating systems \cite{carnieletto2024combining}. In particular, their high thermal inertia makes control inherently complex, often leading to overheating or delayed system responses \cite{qi2024thermal, huang2025experimental}. Conventional non-predictive control strategies, such as proportional-integral (PI) controllers, typically struggle to cope with these dynamic effects \cite{joe2019model}, which may result in increased energy consumption and reduced occupant comfort \cite{kull2017influence}. Consequently, advanced control strategies are required to improve both the responsiveness and the overall performance of UFH systems.

In recent years, machine learning–based approaches have increasingly been explored for the optimization of heating, ventilation, and air conditioning (HVAC) systems \cite{zhou2023comprehensive}. By leveraging data on occupancy patterns, weather conditions, and energy consumption, these methods can predict and optimize thermal behaviour while improving system efficiency \cite{zelazna2025review,kyriakarakos2025artificial}. Among these approaches, Deep Reinforcement Learning (DRL) algorithms have attracted considerable attention \cite{nagy2023ten,al2024reinforcement,wang2020reinforcement,iyanu2024prospects}. Motivated by their success in other complex decision-making domains, DRL methods are particularly well suited for HVAC control due to their ability to operate in dynamic and stochastic environments \cite{singh2022reinforcement,zhao2024survey,yu2021review}. Their adaptability to varying building characteristics, weather conditions, and occupant behaviour enables the simultaneous optimization of energy efficiency and thermal comfort \cite{deng2021reinforcement,coraci2024innovative,manjavacas2024experimental}.

Recent review studies highlight both the opportunities and remaining challenges associated with DRL-based HVAC control \cite{sierla2022review, zhang2022building}, with others specifically demonstrating the potential of these algorithms for managing underfloor heating and cooling systems \cite{han2022model,tang2023indoor,silvestri2024real,rohrer2023deep}. However, this field remains relatively immature, as most studies are conducted in simulated environments. Consequently, the transition to operational buildings is largely unexplored, with few studies addressing the practical requirements for real-world deployment, such as RLOps workflows, system monitoring, and integration with building sensors and actuators \cite{al2024reinforcement,wang2020reinforcement,wang2025simulation,dengiz2024imitation}. As a result, real-world applications remain extremely limited, and commercial DRL controllers are virtually non-existent compared to the widespread use of conventional reactive control methods such as PI controllers, heating curves, and rule-based expert systems \cite{nagy2023ten}.

To address this gap, this work presents the development and evaluation of a commercial DRL-based controller for UFH and HP systems. The study begins with an experimental phase based on EnergyPlus simulations using a pilot single-family house model, where several exploratory analyses are conducted. These include: the evaluation of different DRL control agents, the definition of relevant action and observation variables, the assessment of different control strategies, reward engineering, and the analysis of agent adaptability to dynamic comfort requirements and varying building conditions. 

Based on the insights obtained from the preliminary phase, a case study is then conducted on a real building. This stage involves the development of a digital twin of the building and the subsequent training of DRL agents within the simulation environment. After installing the required equipment in the real building, the best-performing DRL agent is deployed and evaluated for the control of the UFH and HP systems during part of the heating season. The obtained results provide insights into the feasibility and performance of DRL-based control in real operational conditions.

The remainder of this paper is organized as follows. Section \ref{sec:background} presents the theoretical background on DRL and its application to HVAC control. Section \ref{sec:methodology} describes the methodology adopted in this work. Sections \ref{sec:pilot_study} and \ref{sec:case_study} describe the pilot study and the subsequent case study, respectively. Section \ref{sec:discussion} presents and discusses the experimental results and practical considerations. Finally, Section \ref{sec:conclusions} summarizes the main conclusions and outlines directions for future work.


\section{Theoretical background}
\label{sec:background}


\subsection{Deep Reinforcement Learning}
\label{sec:drl}

Reinforcement learning (RL) is a machine learning paradigm based on the interaction between an agent and its environment, typically formalized using Markov Decision Processes (MDPs) \cite{sutton2018reinforcement}. In this framework, an agent interacts with a dynamic process, referred to as the environment, over a discrete sequence of time steps $\mathcal{T} = \{0,1,2,\dots\}$. At each time step~$t \in \mathcal{T}$, the agent observes the current state of the environment~$s_t \in \mathcal{S}$ and selects an action~$a_t \in \mathcal{A}(s_t)$, which influences the evolution of the environment.

In partially observable environments ---formalized as partially observable Markov decision processes (POMDPs)--- the agent does not have direct access to the full state~$s_t$. Instead, it receives an observation~$o_t \in \mathcal{O} \subset \mathcal{S}$ that provides partial information about the underlying state. The relationship between states (or observations) and actions is defined by a policy function $\pi: \mathcal{S} \rightarrow \mathcal{A}$, which represents the agent's decision-making strategy. After executing an action~$a_t$, the environment transitions to a new state~$s_{t+1}$ according to its transition dynamics. At the same time, the agent receives a reward signal~$r_t \in \mathbb{R}$ that evaluates the quality of the transition. This reward signal provides feedback that guides the learning process by updating the policy.

The objective of the agent is to learn an optimal policy~$\pi^*$ that maximizes the expected cumulative reward over time. This objective is commonly expressed through the return $G_t = \sum_{k=0}^{\infty} \gamma^k r_{t+k+1}$, where $\gamma \in [0,1]$ is the discount factor that determines the relative importance of future rewards. The learning process consists of multiple episodes, during which the agent progressively improves its policy based on the rewards received.



The integration of RL with deep neural networks has led to the development of \textit{deep} RL algorithms. In this settings, policies and value functions are approximated using parameterized models, such as neural networks ---i.e., $\pi_{\boldsymbol{\theta}}$ with parameters $\boldsymbol{\theta}$---. DRL algorithms have demonstrated strong capabilities in solving complex optimization and control problems across a wide range of domains \cite{wang2022deep}.


\subsection{HVAC control strategies}
\label{sec:hvac_control}

The primary objective of HVAC control in buildings is to ensure occupant comfort. This objective is commonly defined in terms of thermal comfort, which requires maintaining the indoor temperature within a predefined comfort band. Additional objectives may also be considered, such as minimizing energy consumption or operational costs, controlling humidity levels, or limiting indoor CO$_2$ concentrations \cite{al2024reinforcement}.

Modern HVAC systems provide multiple control interfaces that enable automated operation. These include the adjustment of heating and cooling setpoints across building zones, the regulation of supply water temperature from boilers or heat pumps, the modulation of flow rates through valve positions, and the variation of fan speeds in air handling units (AHUs) \cite{wang2020reinforcement}. Control decisions are typically driven by both endogenous and exogenous variables, such as indoor temperature, humidity, solar radiation, wind conditions, or occupancy-related indicators \cite{zhou2023comprehensive}.

An efficient HVAC control strategy must therefore maintain comfort conditions while minimizing energy consumption. This naturally leads to a multi-objective optimization problem that has been widely studied within the field of building energy management. Several approaches have been proposed to address this problem, although no universally optimal control strategy currently exists. In recent years, DRL algorithms have attracted significant attention due to their potential advantages over conventional approaches such as reactive controllers and Model Predictive Control (MPC) \cite{wang2023comparison,lu2024alternative}.

Reactive control methods remain widely used in industrial settings due to their simplicity and low computational requirements. Typical examples include rule-based strategies, if-else logic, PI controllers, hysteresis controllers, fuzzy control, and expert systems \cite{qi2024thermal}. However, these approaches often rely heavily on expert knowledge and heuristic rules, which results in predefined control strategies with limited adaptability and no predictive capabilities. Consequently, such systems may lead to recurring comfort violations and suboptimal energy performance \cite{nagy2023ten,nguyen2024modelling,pean2019review}.

To overcome these limitations, Model Predictive Control (MPC) has been proposed as a more advanced alternative \cite{drgovna2020all}. MPC computes optimal sequences of control inputs over a finite prediction horizon, typically ranging from several hours to a few days. This approach relies on system models to predict the impact of future control actions on building dynamics. As a result, its performance strongly depends on model accuracy and the ability to account for system uncertainties. Although MPC can achieve near-optimal performance, it requires solving an optimization problem at every control step, which entails significant online computational costs that may limit its practical deployment \cite{wang2023comparison,raman2020reinforcement,arroyo2022reinforced}. In addition, the design and calibration of MPC controllers typically require considerable effort, including the development and tuning of building-specific models \cite{morcego2023reinforcement,brown2023long}.

DRL-based controllers offer a data-driven alternative capable of achieving performance comparable to MPC while requiring minimal online computational effort and providing greater flexibility for deployment \cite{morcego2023reinforcement,jacobs2022reinforcement}. In this paradigm, a control agent learns a policy through iterative interactions with the environment. However, training a control policy directly in real buildings is generally impractical due to the large amount of data and time required. For this reason, agent training is commonly conducted in simulated environments \cite{vazquez2019reinforcement}. In practice, the computational burden is shifted from the online operation phase to an offline training phase, where large amounts of simulated data ---often corresponding to several years of building operation--- are used to learn an effective control policy \cite{blad2022data}.

Offline training typically relies on robust building simulation engines, such as \href{https://energyplus.net/}{EnergyPlus}, \href{https://modelica.org/}{Modelica}, or \href{https://www.mathworks.com/products/simulink.html}{Simulink}, which are used to reproduce building dynamics with sufficient fidelity. In addition, specialized interfaces are required to connect DRL agents with these simulation environments. Several tools have been developed for this purpose, including \href{https://github.com/ugr-sail/sinergym}{Sinergym} \cite{jimenez2021sinergym,campoy2025sinergym} and \href{https://github.com/ibpsa/project1-boptest}{BOPTEST} \cite{blum2021building}.

The performance of DRL-based HVAC controllers depends on several design choices during training. These include the diversity of weather conditions used in simulation, the definition of action and observation spaces, the formulation of the reward function, the control frequency, the selected DRL algorithm, and its associated hyperparameters \cite{manjavacas2024experimental}. In addition, it is important to ensure an appropriate configuration of factors outside the agent's control that nevertheless affect system performance, such as building insulation levels, construction materials, or the sizing and configuration of HVAC equipment.

Despite its promising results, DRL-based HVAC control remains an emerging research area with several open challenges. The number of real-world deployments reported in the literature is still limited, and most practical implementations of advanced building control strategies are based on MPC \cite{wang2024long}. Scalability also remains an open question, particularly regarding the potential use of transfer learning to generalize control policies trained on a limited number of buildings to other building types \cite{wang2020reinforcement}. Another critical issue is explainability \cite{milani2024explainable,weinberg2023review,jimenez2023explaining}. Since DRL policies are often treated as black-box models, additional analysis may be required to interpret agent decisions and ensure the safety and reliability of the controlled system.

Based on the real-world DRL implementation challenges outlined in \cite{dulac2021challenges}, HVAC control presents several characteristics that make it a challenging application domain for DRL. The environment is stochastic and partially observable, with a high-dimensional and continuous state space. Safety constraints must be respected, and explainability is required for both system operators and end users. In addition, the agent must optimize a multi-objective reward function typically composed of several sub-rewards \cite{togashi2025reward}. Finally, practical deployment involves strict timing requirements for inference (often on the order of minutes), as well as potential delays in sensors and actuators.


\subsection{DRL applications in underfloor settings}
\label{sec:drl_underfloor}

While most DRL applications in HVAC control focus on conventional air-based systems, radiant heating and cooling technologies have received comparatively less attention \cite{al2024reinforcement,arroyo2022reinforced}. These systems are widely adopted in energy-efficient buildings due to their ability to provide uniform thermal comfort and efficient thermal distribution. However, their dynamic behaviour differs significantly from that of air-based systems, primarily because of their high thermal inertia and delayed response to control actions. These characteristics increase the difficulty of designing effective control strategies and make reactive approaches suboptimal, highlighting the potential benefits of predictive and learning-based methods such as DRL.

Several studies have evaluated the potential of DRL to improve the energy performance of UFH systems in simulation environments. For instance, \cite{han2022model} compares multiple DRL approaches with a rule-based controller (RBC) that incorporates weather forecasting. The results show a 19.1\% reduction in heating demand using DRL with weather forecasts, compared to an 18.5\% reduction achieved with MPC. Similarly, \cite{tang2023indoor} demonstrates that a DRL controller can maintain indoor comfort while reducing energy consumption by 3.19\% to 6.30\% compared to an on–off controller, and by up to 10.48\% relative to a PI-based variable water temperature control.

Other works have analyzed the robustness and adaptability of DRL controllers under varying operating conditions. For example, \cite{brandi2022energy} evaluates the performance of DRL in controlling a UFH system compared to an RBC based on indoor temperature and occupancy schedules. Multiple simulation scenarios are considered, including variations in outdoor conditions, temperature requirements, and building characteristics. The results indicate energy savings of approximately 6\% while maintaining indoor comfort requirements.

In \cite{rohrer2023deep}, DRL is used to control a heat pump combined with a UFH system in buildings with different thermal characteristics. The results show that DRL can achieve demand-response performance comparable to MPC, although some comfort deviations may occur. Similarly, \cite{silvestri2024real} investigates the intelligent control of a thermally activated building system (TABS) in a real building. In this study, a surrogate resistance–capacitance model is used as a training environment to compare rule-based, PI, MPC, and DRL controllers. The algorithms are then deployed in the real building during the cooling season to regulate the supply temperature of the TABS, showing strong adaptability of the DRL controller and improved performance compared to reactive approaches.

In addition, \cite{blad2023laboratory} presents a laboratory implementation of DRL applied to a real UFH system. This work combines offline DRL training with experimental validation, demonstrating stable control behaviour, predictive-like operation, and reduced temperature oscillations. Among the actuators to be controlled are the supply water temperature and valve positions for each controlled thermal zone. 

Despite these promising results, several studies highlight the limitations of DRL for radiant systems control. For instance, \cite{blad2019control} evaluates a DRL controller under different system configurations, varying the number of zones and the thermo-physical properties of the building. The results indicate that more complex configurations, such as multi-zone buildings, are significantly harder for the agent to control effectively. To address these limitations, hybrid control strategies have been proposed. For example, \cite{arroyo2022reinforced} introduces a hybrid MPC–DRL approach for a simulated underfloor cooling (UFC) system, showing how the complementary strengths of both methods can improve overall control performance.

More advanced DRL configurations have also been investigated. Multi-agent approaches have been proposed to address coupled control problems, such as the joint control of UFC and dehumidification \cite{liu2024joint}.  Finally, recent research has explored the integration of physics-informed learning techniques within DRL frameworks. In \cite{wang2025physics}, a DRL controller based on a physics-consistent input convex neural network is deployed in a real radiant ceiling system. The proposed approach achieves a 33\% reduction in panel loads compared to a baseline controller, highlighting the potential of combining physical knowledge with data-driven control methods.

\section{Methodology}
\label{sec:methodology}

\begin{figure}
    \centering
    \includegraphics[width=\linewidth]{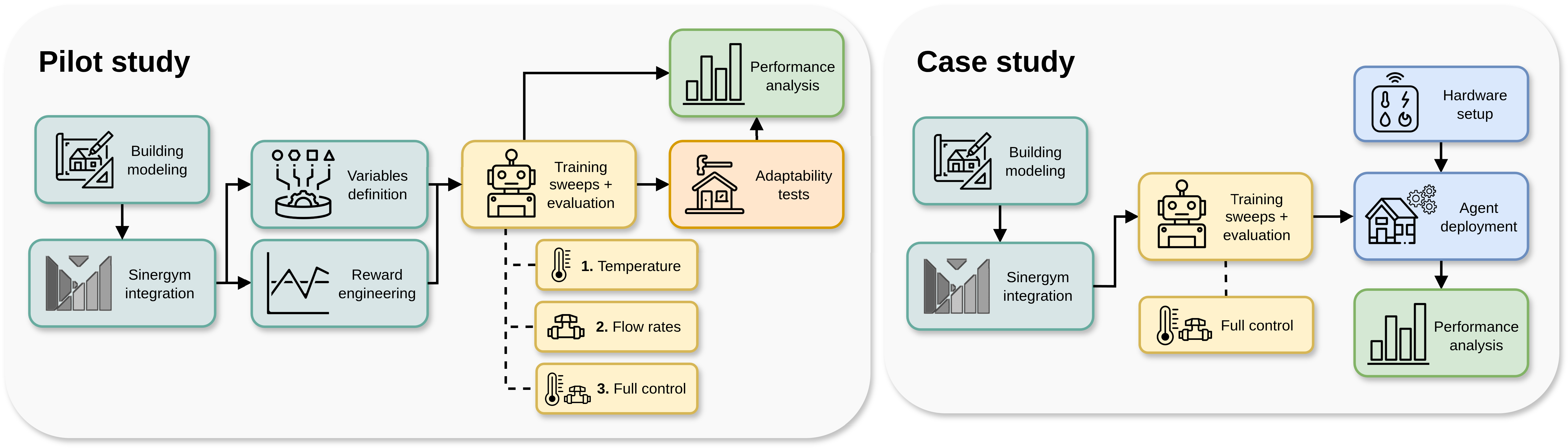}
    \caption{Overview of the proposed two-phase methodology used to design, train, and validate the DRL controller in simulation and real-building operation}
    \label{fig:methodology}
\end{figure}

The main objective of this work is to address the practical deployment of DRL controllers for UFH systems coupled with HPs, and to evaluate their performance in a real residential building. Starting from simulation environments where control agents are trained and evaluated, a series of experiments are conducted to support the subsequent deployment of the controller in an occupied dwelling. As illustrated in Figure~\ref{fig:methodology}, the experimental procedure is structured into two main phases:

\begin{itemize}
    \item The first phase consists of a \textit{pilot study} carried out on a generic simulated building to assess the performance of different DRL agents, control strategies, and environment configurations. The insights gained from this exploratory stage are then used to inform the implementation of a DRL agent in a concrete real building.
    \item The second phase comprises the \textit{case study}, with the development of a real-building digital twin for simulation-based training, followed by the deployment and monitoring of the trained agent under real operating conditions.
\end{itemize} 

The following subsections provide a detailed description of the problem formulation and the experimental setup used in both studies.


\subsection{Problem formalization}
\label{sec:formalization}

The control problem is defined as a multi-objective optimization task aimed at maintaining thermal comfort while minimizing energy consumption. The problem can be formulated as a POMDP composed of the following elements:


\paragraph{Observation space}

Each observation $o_t \in \mathcal{O}$ is represented as a vector of $M$ features $(o^1_t, o^2_t, \dots, o^M_t) \in \mathbb{R}^{M}$ containing the information perceived by the agent at time step $t \in \mathbb N$ to determine the next control action. These observations include exogenous variables (e.g., outdoor temperature, humidity, wind conditions, solar radiation), endogenous variables (e.g., indoor temperatures and humidity), and time-related variables (e.g., day of the year, month, and hour).


\paragraph{Action space}

Based on the information contained in each observation, the policy outputs an action $a_t \in \mathcal{A}$ whose definition depends on the adopted control strategy. We distinguish three control strategies, hereafter referred to as cases:

\begin{itemize}

\item \textit{Case~1}. Each action $a_t \in \mathbb{R}$ corresponds to the heat pump supply water temperature, bounded within $[25, 45]$~\textcelsius. Flow rate control is not performed by the DRL agent; instead, it is adjusted proportionally according to the thermal demand of each circuit.

\item \textit{Case~2}. Given $N$ controlled circuits, the action $a_t \in \mathbb{R}^{N}$ is defined as a vector $a_t = (a^1_t, a^2_t, \dots, a^N_t) \in [0,1]^N$, representing the valve opening fraction of each UFH circuit. Although valve openings are defined as continuous values, they are implemented as binary positions in practice, where 0 represents a fully closed valve and 1 a fully open valve. In this case, the heat pump supply temperature remains constant at 35~\textcelsius.

\item \textit{Case~3}. Each action $a_t \in \mathbb{R}^{N+1}$ jointly controls valve openings and supply temperature, such that 
\[
a_t = ((a^1_t, a^2_t, \dots, a^N_t), a^{N+1}_t),
\]
where $(a^1_t, a^2_t, \dots, a^N_t) \in [0,1]^N$ represent the valve openings and $a^{N+1}_t \in \mathbb{R}$ corresponds to the supply water temperature.

\end{itemize}


\paragraph{Reward function}

Given the multi-objective nature of the problem, the reward signal obtained at each time step $r_t \in \mathbb{R}$ is defined as a weighted linear combination of comfort and energy consumption terms:

\begin{equation}
\label{eq:reward}
r_t = - \omega \lambda_P \frac{P_t}{\sqrt{1+\bar{C}_t}} - (1 - \omega)\lambda_C \bar{C}_t,
\end{equation}

where $\omega \in [0,1]$ determines the relative importance assigned to comfort and energy consumption; $\lambda_P$ and $\lambda_C$ are scaling factors used to balance the magnitude of the two terms; $P_t$ denotes the HP power consumption (in W); and $\bar{C}_t$ represents the mean comfort violation across all building zones.

Given a building divided into $N$ zones, the mean comfort violation is defined as

\begin{equation}
    \bar{C}_t =
    \frac{1}{N}\sum_{i=1}^{N}
    \Big(
    \max(T^i_{\text{low},t}-T^i_t,0)
    +
    \max(T^i_t-T^i_{\text{up},t},0)
    \Big),
\end{equation}

where $T^i_t$ (in \textcelsius) denotes the indoor temperature of zone $i$, and $T^i_{\text{low},t}$ and $T^i_{\text{up},t}$ represent the lower and upper comfort bounds, respectively. This formulation allows the agent to allocate additional energy when comfort violations occur, thereby prioritizing comfort while treating energy consumption as a secondary objective under discomfort conditions \cite{brandi2020deep}. Moreover, the formulation allows different zones to have distinct comfort limits that may vary over time.


\paragraph{Transition function}

State transitions are governed by a transition function $f:\mathcal{S}\times\mathcal{A}\rightarrow\mathcal{S}$. In real buildings this function is inherently unknown and determined by complex, nonlinear thermal dynamics. In the simulation environment, these dynamics are emulated by a simulation engine ---i.e., EnergyPlus--- which computes the resulting state transitions based on a predefined model of the building.


\subsection{Experimental setup}
\label{sec:exp_setup}

\begin{figure}
    \centering
    \includegraphics[width=.4\linewidth]{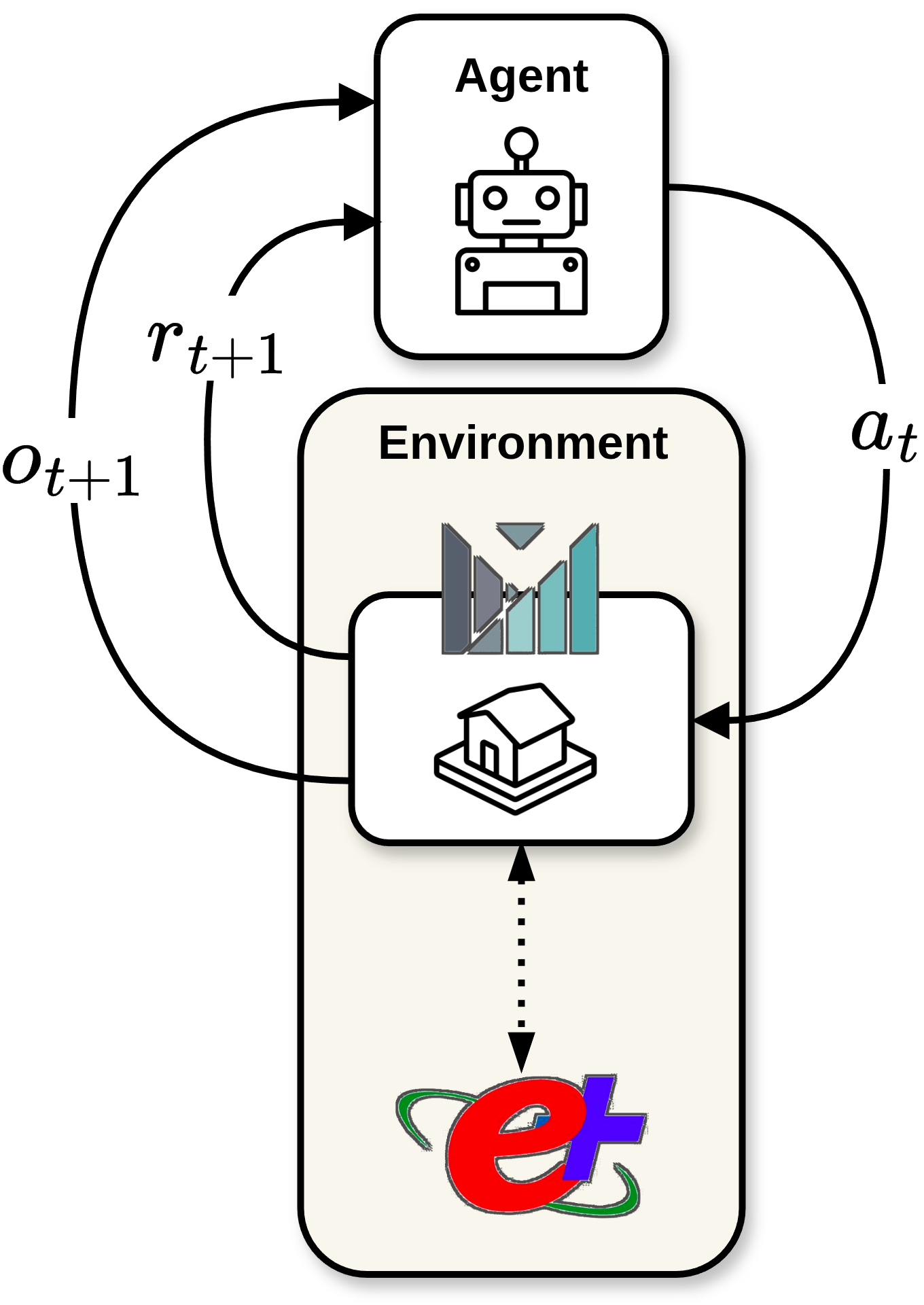}
    \caption{Overview of the agent–environment interaction during simulation-based training and evaluation, using Sinergym as the interface between the DRL agent and EnergyPlus}
    \label{fig:mdp}
\end{figure}

All simulation experiments are conducted using EnergyPlus as the building simulation engine. As shown in \autoref{fig:mdp}, the connection between DRL agents and EnergyPlus simulations is established through the Sinergym framework \cite{jimenez2021sinergym,campoy2025sinergym}, which have been widely used in the literature and provides multiple tools to facilitate training and evaluation of DRL agents \cite{campoy2026sinergym}. Internal communication between Sinergym and the building simulation is handled via the EnergyPlus Python API.

Each simulated episode represents a complete heating season, spanning from 15 November to 15 March. Agent–environment interactions occur every 30 minutes. This interval was selected after extensive testing of different control actuation frequencies, considering both the heat pump compressor cycling frequency and the resulting control performance. It allows the agent to observe the effects of its actions while avoiding potential degradation of the HP due to short cycling.

To expose the agent to a diverse range of operating conditions during training, both environmental and occupant-related variables are randomized across episodes. Weather conditions are based on EnergyPlus weather datasets (EPW) corresponding to the expected building location, with Ornstein–Uhlenbeck noise applied to the most influential weather variables (outdoor temperature, humidity, wind, and solar radiation) across episodes. This procedure introduces variability while preserving annual trends, enhancing agent robustness and reducing the risk of overfitting. To emulate user interaction, random setpoint variations in $[19, 21]$\textcelsius{} are applied at each timestep for each building zone with a 5\% probability, maintaining a comfort band of $\pm1$\textcelsius{} around the setpoint.

Regarding agents, implementations of several state-of-the-art DRL algorithms from the \href{https://github.com/DLR-RM/stable-baselines3}{Stable-Baselines3} library are employed, including Soft Actor-Critic (\texttt{SAC}) \cite{haarnoja2018soft}, Truncated Quantile Critics (\texttt{TQC}) \cite{kuznetsov2020controlling}, Proximal Policy Optimization (\texttt{PPO}) \cite{schulman2017proximal}, and \texttt{PPO} with a recurrent policy (\texttt{RecPPO}) \cite{raffin2021stable}. Hyperparameter tuning and model optimization are conducted via grid search sweeps using \href{https://wandb.ai}{Weights \& Biases}. Agents are trained until reward convergence is achieved (100 episodes for the pilot study, and 200 episodes for the case study), with periodic evaluations conducted every 10 episodes to select the best-performing agent during training.

Once trained, the DRL agents are evaluated over 10 episodes using different random seeds. Performance is assessed using multiple metrics, including overall statistics (mean episode reward, average comfort violations, mean energy consumption) and episode-specific measures, such as the distribution of HP supply water temperatures and the valve openings of each UFH circuit.


\section{Pilot study}
\label{sec:pilot_study}

As a starting point, a pilot study was conducted to evaluate the control capabilities of DRL algorithms in a simulated building environment. The goal was not to compare DRL against other controllers, but rather to identify the strategies that allow DRL to achieve satisfactory performance, as well as the technical and formal requirements for its implementation. As already indicated in the previous section, different HP and UFH control strategies, agents, and configurations were considered in this initial phase. The study also established the workflow and infrastructure required for simulation-based training and agent evaluation.

\begin{figure}
    \centering
    \includegraphics[width=.85\linewidth]{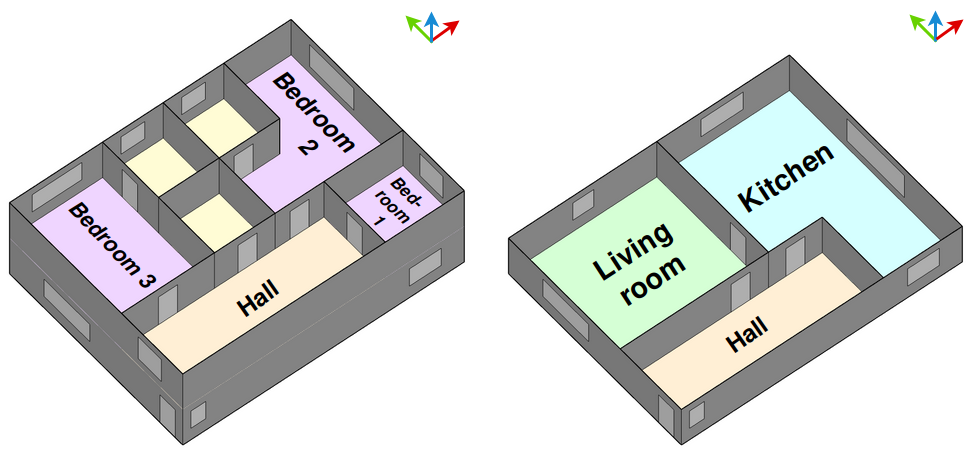}
    \caption{Thermal zoning and spatial layout of the pilot building modeled in EnergyPlus for the simulation-based evaluation of the control strategies}
    \label{fig:pilot_building}
\end{figure}

An EnergyPlus \href{https://github.com/ugr-sail/sinergym/blob/uponor-develop/sinergym/data/buildings/radiant_residential_building_hp_heating.epJSON}{simulation model} of a semi-detached single-family home is used as the reference building during this phase. As shown in \autoref{fig:pilot_building}, the building consists of two floors and is divided into five main zones: a living room, a kitchen, and three bedrooms. The building is simulated using the climatic conditions of Stockholm (mean annual temperature: $6.43^\circ$C; mean relative humidity: $78.42$\%), representing a cold climate that poses a demanding control scenario for the agent. The building HVAC system is equipped with an air-to-water HP supplying UFH, with a single circuit per room and individualized flow rate control. The complete list of observed variables considered by the agents is provided in \ref{app:variables}. 

\begin{figure}
    \centering
    \includegraphics[width=\linewidth]{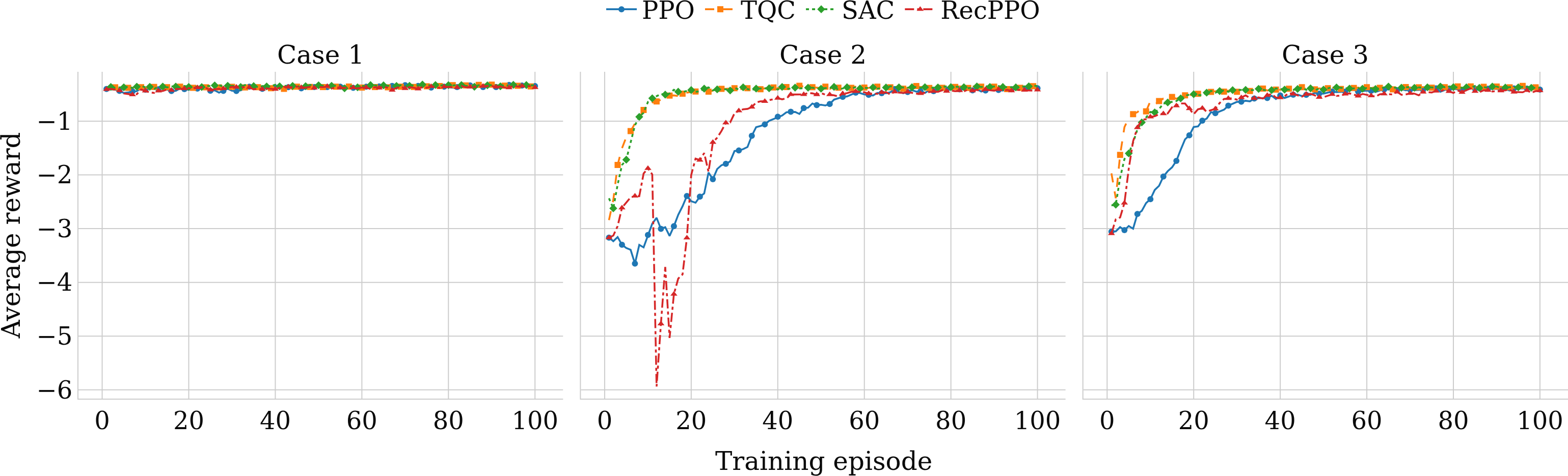}
    \caption{Reward evolution during training for the different control cases and DRL agents evaluated in the pilot study}
    \label{fig:reward_evolution}
\end{figure}

For training, a reward function with $\omega = 0.5$ was employed (see \autoref{eq:reward}), assigning equal weight to comfort and energy consumption, while variable setpoints were considered during both training and evaluation. Figure~\ref{fig:reward_evolution} illustrates the reward evolution across the different cases and DRL agents. As expected, Cases 2 and 3 exhibit slower convergence, likely due to the increased problem complexity associated with their larger action spaces. Overall, \texttt{SAC} and \texttt{TQC} exhibit the fastest convergence rates, whereas \texttt{PPO} and \texttt{RecPPO}, owing to their on-policy nature, require substantially more training data to achieve comparable performance. Furthermore, \texttt{RecPPO} generally displays greater instability during the early stages of training.

\begin{figure}
    \centering
    \includegraphics[width=\linewidth]{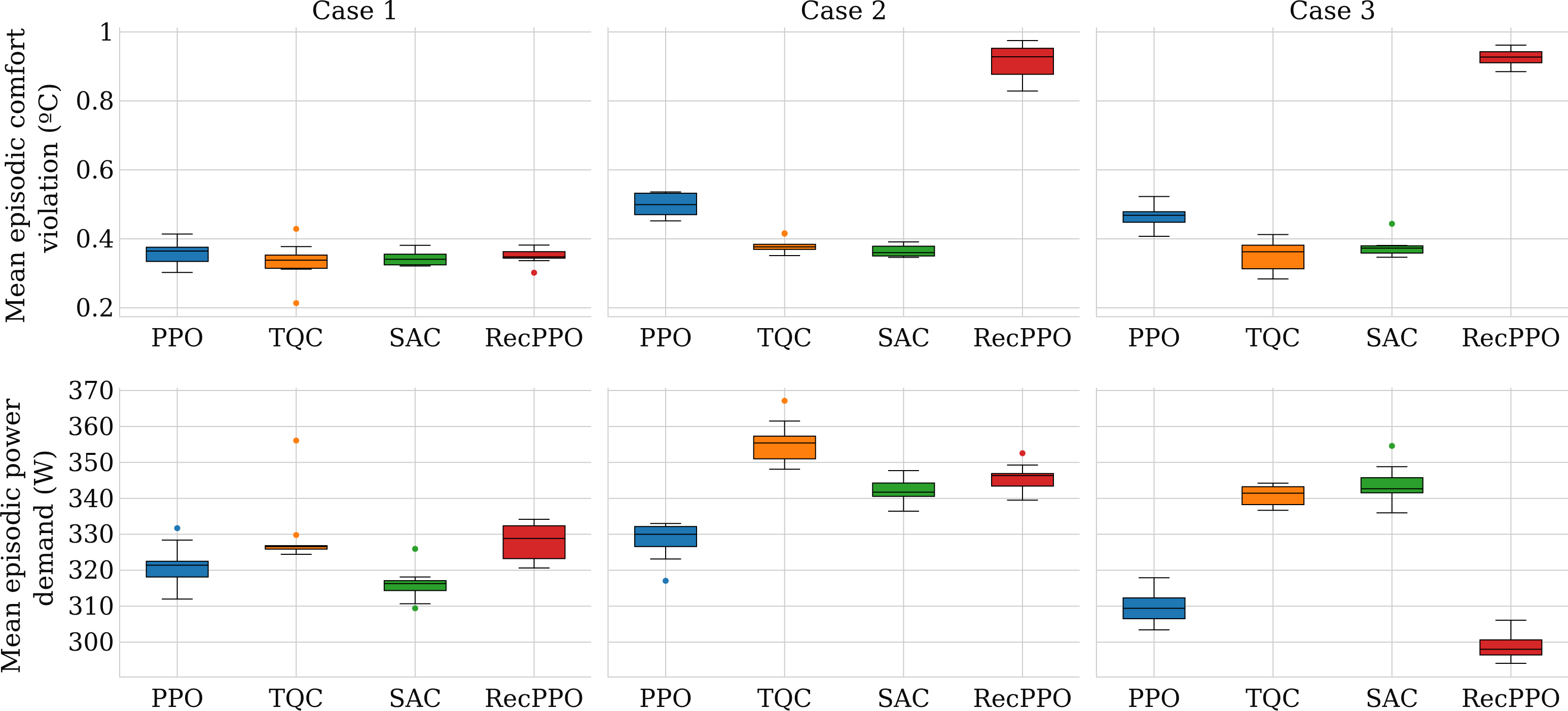}
    \caption{Mean episodic comfort violations and power demand for different control cases and DRL agents in the pilot study}
    \label{fig:comfort_power_boxplots}
\end{figure}

Once training was completed, the agents were evaluated. Figure~\ref{fig:comfort_power_boxplots} presents boxplots of the mean episodic comfort violation and power demand for each agent-case combination. The results reveal greater variability in performance under the more complex control scenarios (Cases 2 and 3), reflecting the increased difficulty of the learning task. A quantitative summary of the evaluation metrics is provided in Table~\ref{tab:pilot_study_metrics}, which shows that \texttt{SAC} achieves the highest mean episodic reward for Cases 1 and 2, while \texttt{TQC} is the best performing agent for Case~3.

Although Cases 1 and 2 achieve performance levels comparable to those of Case~3, the latter is selected as the default control strategy. This choice is motivated by the main objective of the work, namely, to investigate the capabilities of DRL agents in environments where they are granted full, or at least the highest possible degree of, control over the system. Case~1 relies on an automatic valve control scheme implemented by the simulator, corresponding to a proportional control mechanism that is internally solved by EnergyPlus and is not expected to be available in the real-building. In contrast, Case~2 restricts the agent's actions to valve control only, thereby limiting its ability to achieve energy savings through the use of lower supply water temperatures. Consequently, Case~3 provides the most suitable framework for assessing the potential of DRL-based control under minimal external control assumptions. Based on these results, the \texttt{TQC} agent trained under Case~3 is adopted as the baseline agent for the remainder of the study.

\begin{table}[ht]
    \centering
    \footnotesize
    \begin{tabular}{c c l l l}
        \toprule
        {} & {} & \textbf{Comfort violation (ºC)} & \textbf{Power consumption (W)} & \textbf{Reward} \\
        \midrule

        \multirow{4}{*}{Case~1} & \texttt{PPO}   & $0.34 \pm 0.92$ & $317.39 \pm 202.57$ & $-0.32 \pm 0.43$ \\
                               & \texttt{TQC}    & $0.42 \pm 1.25$ & $326.56 \pm 226.16$ & $-0.37 \pm 0.59$ \\
                               & \texttt{SAC}    & $\mathbf{0.32 \pm 0.89}$ & $\mathbf{317.09 \pm 215.04}$ & $\mathbf{-0.31 \pm 0.42}$ \\
                               & \texttt{RecPPO} & $0.35 \pm 0.88$ & $321.20 \pm 267.61$ & $-0.33 \pm 0.42$ \\
        \midrule
        \multirow{4}{*}{Case~2} & \texttt{PPO}   & $0.47 \pm 1.07$ & $\mathbf{331.01 \pm 318.91}$ & $-0.39 \pm 0.53$ \\
                               & \texttt{TQC}    & $0.41 \pm 1.31$ & $357.31 \pm 378.87$ & $-0.38 \pm 0.64$ \\
                               & \texttt{SAC}    & $\mathbf{0.39 \pm 1.25}$ & $336.42 \pm 341.19$ & $\mathbf{-0.36 \pm 0.61}$ \\
                               & \texttt{RecPPO} & $0.87 \pm 1.16$ & $346.49 \pm 412.61$ & $-0.59 \pm 0.61$ \\
        \midrule
        \multirow{4}{*}{Case~3} & \texttt{PPO}   & $0.47 \pm 1.25$ & $310.20 \pm 228.16$ & $-0.38 \pm 0.60$ \\
                               & \texttt{TQC}    & $\mathbf{0.30 \pm 1.09}$ & $343.25 \pm 326.25$ & $\mathbf{-0.32 \pm 0.53}$ \\
                               & \texttt{SAC}    & $0.37 \pm 1.13$ & $354.59 \pm 381.42$ & $-0.36 \pm 0.56$ \\
                               & \texttt{RecPPO} & $0.91 \pm 0.99$ & $\mathbf{298.63 \pm 23.23}$ & $-0.59 \pm 0.49$ \\

        \bottomrule
    \end{tabular}
    \caption{Summary of performance metrics obtained from the pilot study evaluation (episodic mean $\pm$ standard deviation)}
    \label{tab:pilot_study_metrics}
\end{table}

\begin{figure}
    \centering
    \includegraphics[width=\linewidth]{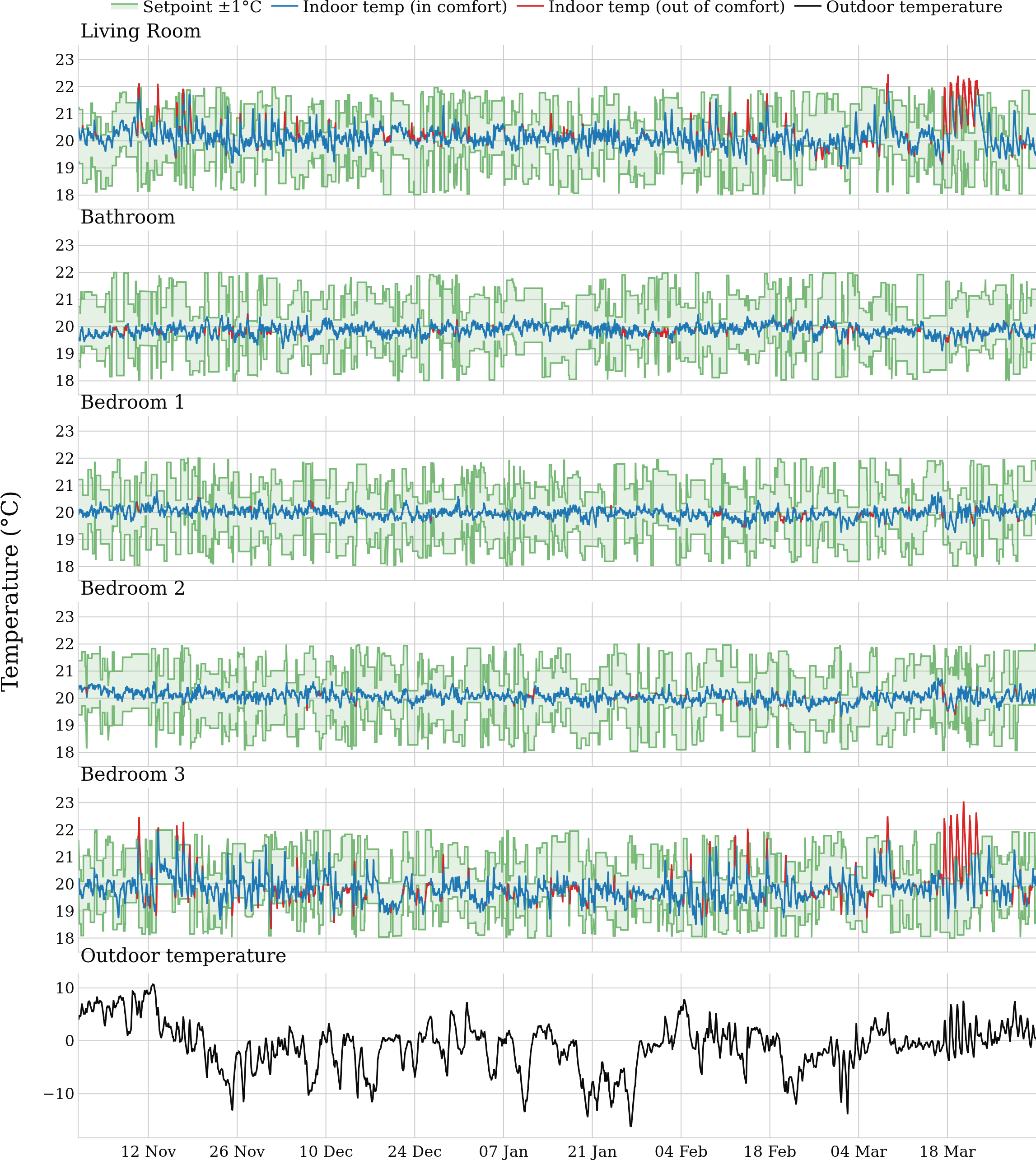}
    \caption{Indoor and outdoor temperatures evolution during a complete heating season in the pilot building (\texttt{TQC} - Case~3)}
    \label{fig:best_agent_evolution}
\end{figure}

A comprehensive performance analysis of this baseline agent was subsequently conducted. Figure~\ref{fig:best_agent_evolution} illustrates the evolution of the indoor temperatures in each room throughout the heating season. Despite the introduction of random setpoint variations, the indoor temperatures remain within the comfort range for the majority of the evaluation period, indicating that the agent is able to maintain satisfactory thermal comfort under varying operating conditions. The largest comfort violations are observed at the beginning and end of the heating season (e.g., in \textit{Bedroom~3}), where the increased influence of solar gains makes temperature regulation more challenging. It should be noted, however, that the heating season considered in this study has a fixed duration for reproducibility purposes. In practice, the start and end of the heating season are typically determined by the occupant according to weather conditions, meaning that these transition periods, and the associated control challenges, may be shorter or may not occur at all in a real deployment.

Having established the baseline agent's performance in the nominal building configuration, the next step was to evaluate its robustness under a range of degraded building conditions. This analysis is particularly relevant in the context of real-world deployment, where discrepancies between the simulation model and the physical building are unavoidable. Assessing the impact of such discrepancies provides insight into the performance degradation that can be expected when transferring the controller from simulation to a real environment.

\begin{figure}
    \centering
    \begin{subfigure}{\linewidth}
        \centering
        \includegraphics[width=\linewidth]{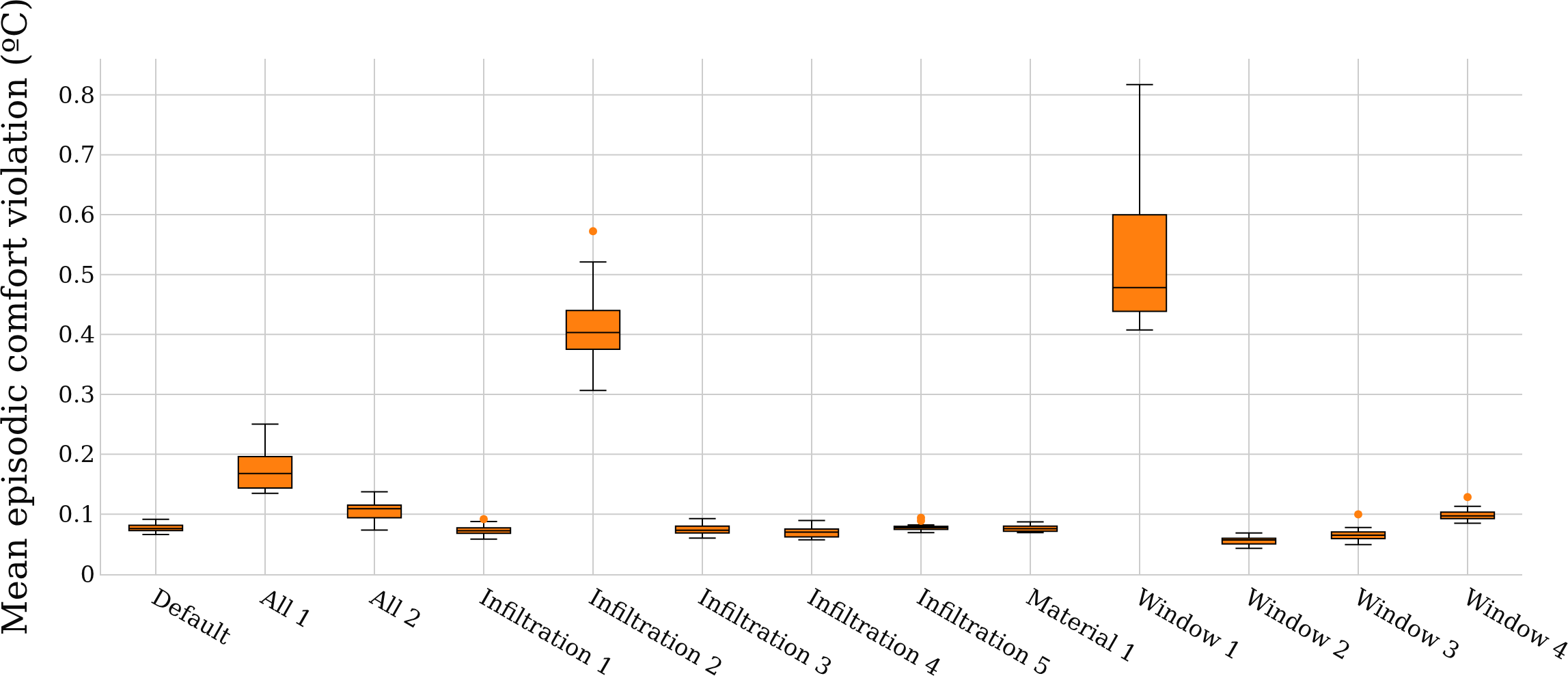}
        \caption{Comfort violations}
        \label{fig:degradations_1}
    \end{subfigure}

    \vspace{0.5em}

    \begin{subfigure}{\linewidth}
        \centering
        \includegraphics[width=\linewidth]{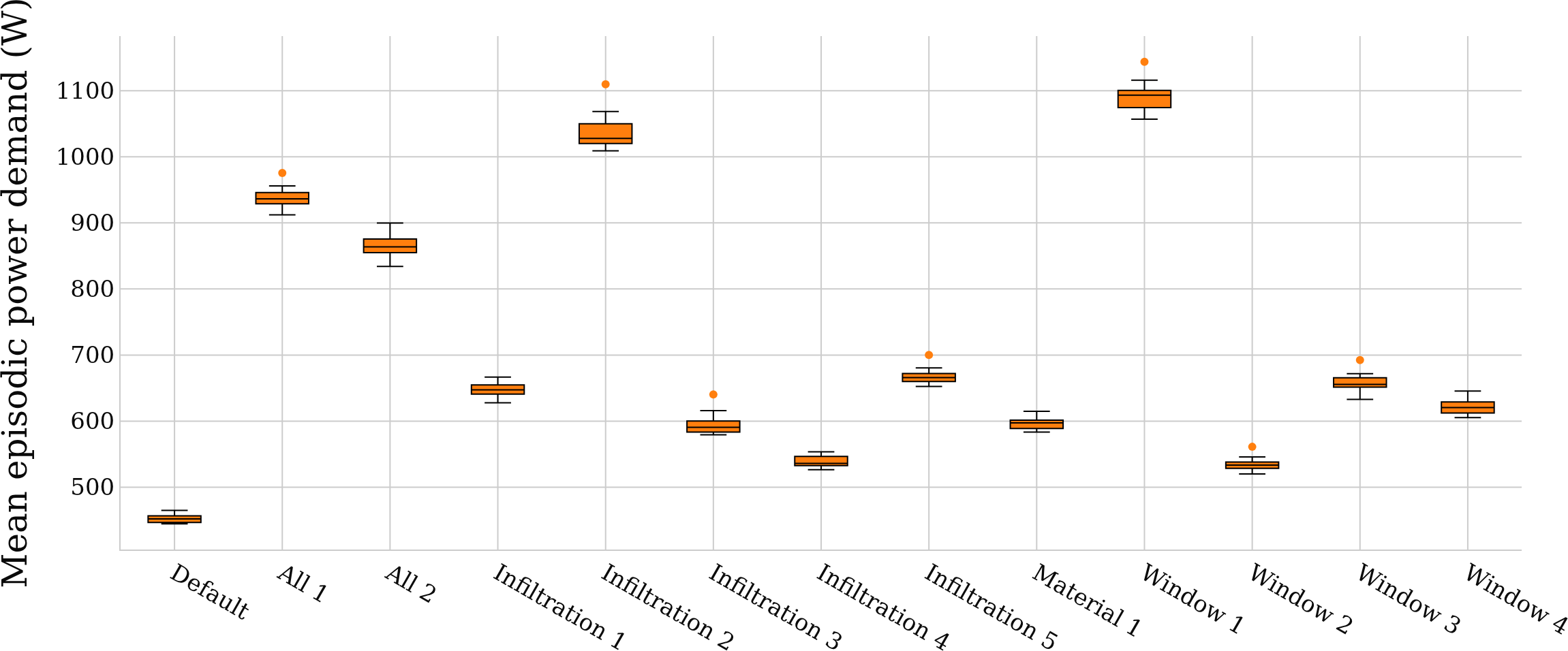}
        \caption{Power demand}
        \label{fig:degradations_2}
    \end{subfigure}

    \caption{Performance under different degradation scenarios compared with the default pilot building configuration (\texttt{TQC} -- Case~3). The robustness tests include variations in the building envelope, considering different wall insulation levels (\textit{Infiltration}), window characteristics (\textit{Window}), material properties (\textit{Material}), and combinations of all these modifications (\textit{All})}
        \label{fig:degradations}
\end{figure}

To this end, the baseline agent was evaluated under a series of alternative building configurations designed to represent previously unseen operating conditions. These configurations, described in \ref{app:degradations}, introduce variations (i.e., degradations) in building properties. As shown in Figure~\ref{fig:degradations}, the agent maintains thermal comfort across all tested scenarios, with average comfort violations remaining close to $0.1^\circ$C in most cases and only a few configurations reaching approximately $0.2^\circ$C, $0.4^\circ$C, and $0.5^\circ$C. This robustness, however, comes at the cost of increased energy consumption in all degraded scenarios. Overall, these results suggest that the learned policy is able to adapt to changes in the building's thermal characteristics, preserving comfort despite modeling inaccuracies and variations in the building envelope.

In conclusion, the pilot study yielded several important insights, which can be summarized as follows:

\begin{itemize}
    \item Off-policy DRL agents (\texttt{SAC} and \texttt{TQC}) exhibit superior performance, stability, and learning efficiency compared with on-policy agents (\texttt{PPO} and \texttt{RecPPO}).

    \item The \texttt{TQC} agent achieved the best overall performance for the most complex control scenario (Case~3), and demonstrated the ability to maintain thermal comfort throughout most of the heating season under varying setpoint conditions.
    
    \item Full control of both the HP supply temperature and UFH valves (Case~3) was selected as the preferred strategy, as it maximizes the agent's control capabilities while avoiding reliance on external control mechanisms or predefined supply temperature schedules.
    
    \item The baseline agent exhibited strong generalization capabilities, maintaining low comfort violations (typically around 0.1°C) when evaluated in previously unseen building configurations with degraded insulation properties. However, the agent's adaptation to previously unseen building properties is associated with increased energy consumption.
\end{itemize}

Overall, this experimental phase made it possible to identify the workflow, system configuration, and key challenges associated with DRL training and evaluation. These findings provided the foundation for the subsequent case study and informed the design decisions adopted for the deployment in a real building environment.


\section{Case study}
\label{sec:case_study}

Following the initial pilot study, a second and considerably more challenging case study was developed to assess the scalability of the proposed DRL-based control strategy in a real-world residential setting. The case study considers an occupied detached dwelling that was specifically instrumented to enable experimental validation under real operating conditions. Within this framework, two main objectives were pursued: (i) evaluating the ability of DRL agents to control a substantially more complex building, and (ii) validating the performance of a DRL agent trained in simulation and subsequently deployed in the real building.

\begin{figure}
    \centering

    \begin{subfigure}{\linewidth}
        \centering
        \includegraphics[width=\linewidth]{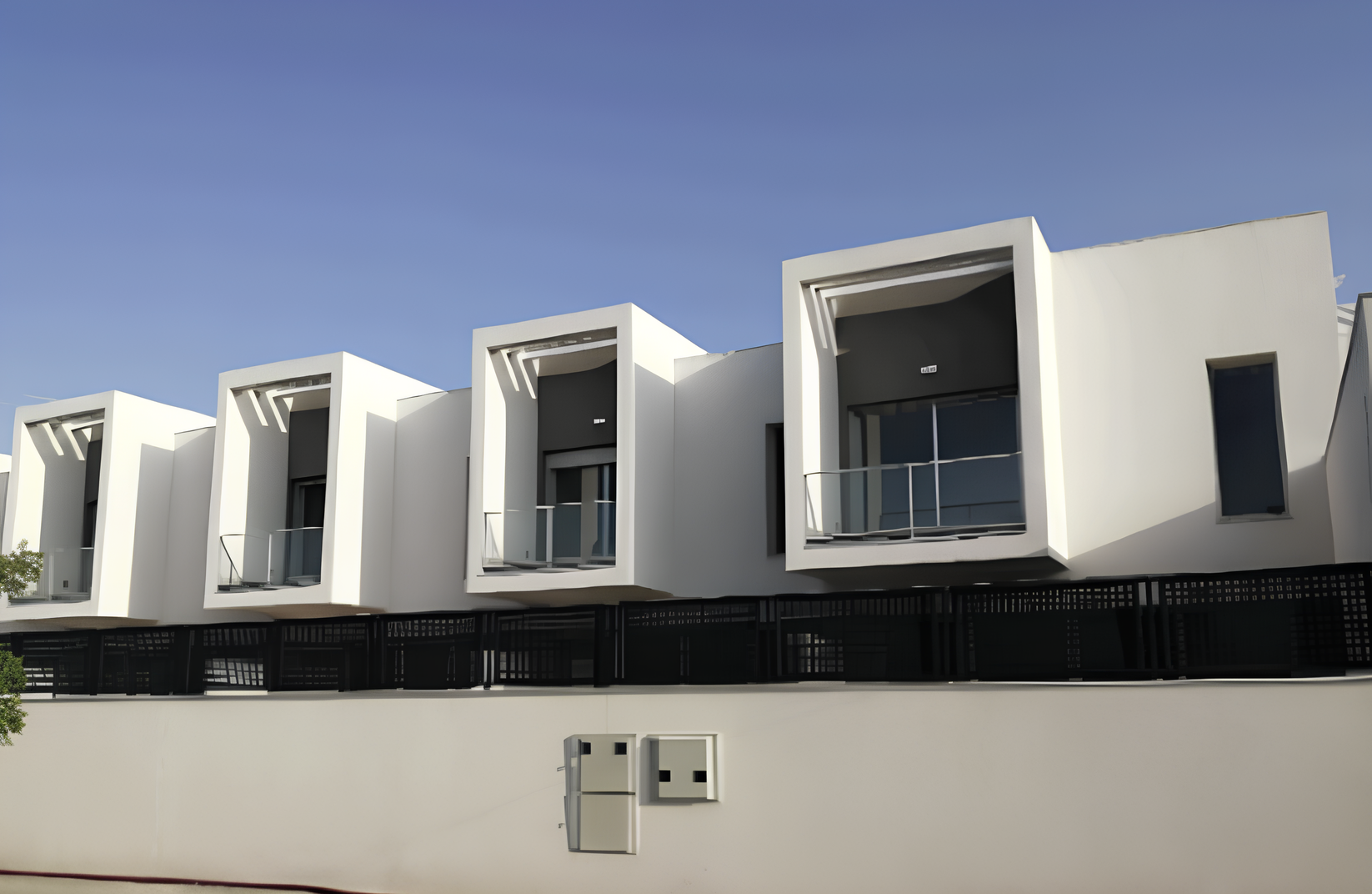}
        \caption{Real building image}
        \label{fig:case_study_building_real}
    \end{subfigure}

    \vspace{0.5em}

    \begin{subfigure}{\linewidth}
        \centering
        \includegraphics[width=\linewidth]{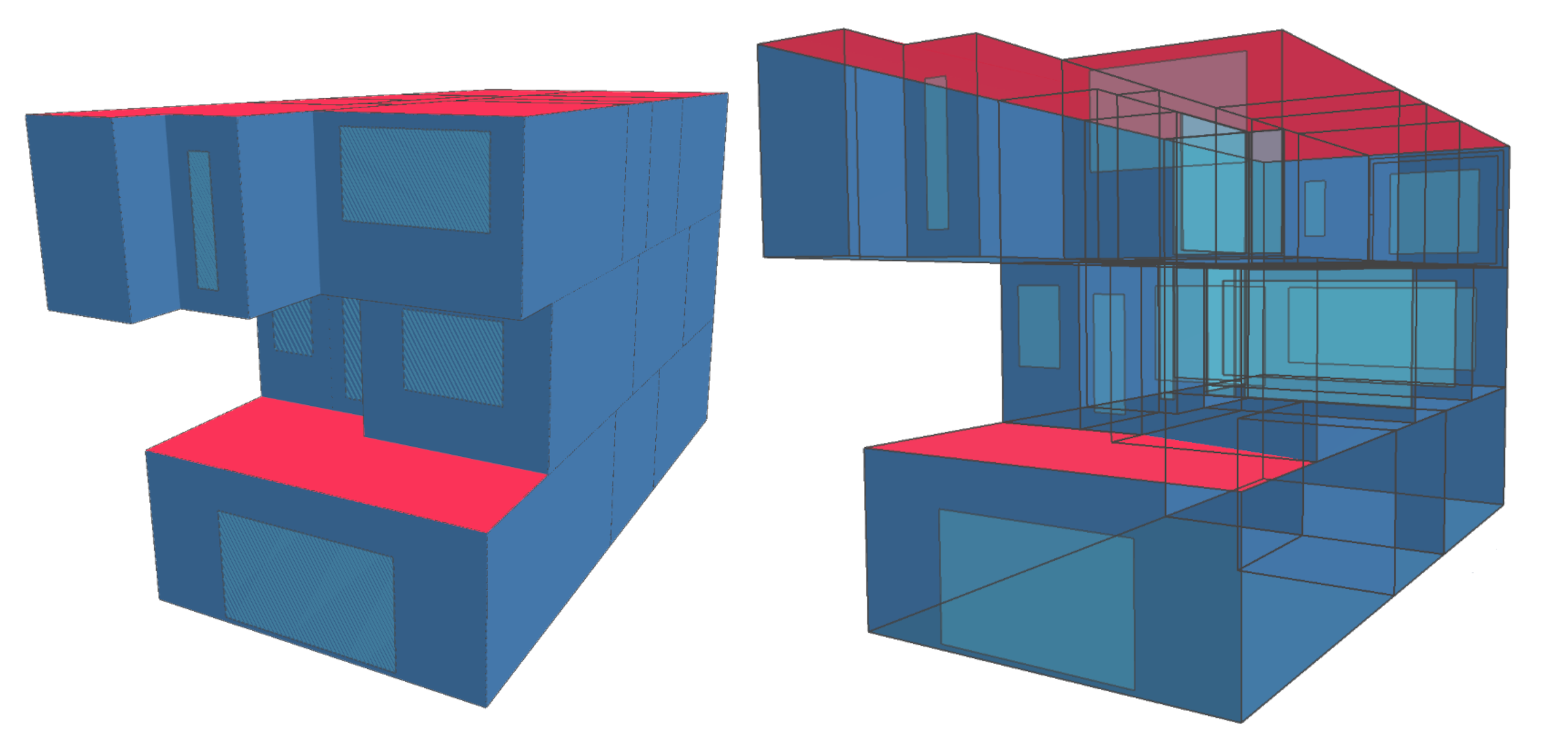}
        \caption{EnergyPlus model of the building}
        \label{fig:case_study_building_model}
    \end{subfigure}

    \caption{Residential building considered in the case study and its corresponding digital twin. The dwelling consists of a basement, a ground floor, and a first floor, comprising thirteen independently controlled thermal zones}
    \label{fig:case_study_building}
\end{figure}

For this study, the building is considerably larger than the one considered in the pilot study and comprises three floors interconnected by a central staircase: a basement, a ground floor, and a first floor, as illustrated in Figure~\ref{fig:case_study_building}. The basement contains a storage area, a warehouse and the technical room where the heating equipment is installed. The ground floor comprises the main living spaces, including kitchen-living room, hallway and bathroom. The first floor contains three bedrooms, two bathrooms, a walk-in closet and an attic area. In total, the building is divided into seven thermal zones, each one equipped with an independent UFH circuit connected to the Uponor \textit{Smatrix}\textsuperscript{\textcopyright} control system.

As in the pilot study, the heating system consists of an air-to-water HP supplying the UFH installation. The DRL controller directly determines both the HP supply water temperature and the opening position of every UFH valve, following the same control strategy selected in Case~3 of the pilot study. However, the dimensionality of the control problem increases due to the larger number of thermal zones and hydraulic circuits.

To accelerate the training process, an EnergyPlus digital twin of the dwelling was developed. The model reproduces the building geometry, construction materials, thermal zoning, hydronic UFH system, and occupancy schedules, enabling the DRL agents to be trained in simulation before deployment in the real building. Simulations are performed using the climatic conditions of Madrid (mean annual temperature: $14.3$ ºC; mean relative humidity: $61.6$\%), where the real building is located. 

Unlike the pilot study, which primarily focused on comparing different DRL algorithms and control strategies, this case study aims to evaluate the complete deployment workflow. The study therefore begins with the training and evaluation of different DRL agents using the digital twin, followed by the deployment of the selected controller in the real dwelling and a comparison between simulated and real-world operation.

\subsection{Training in the digital twin}

\begin{figure}
    \centering
    \includegraphics[width=\linewidth]{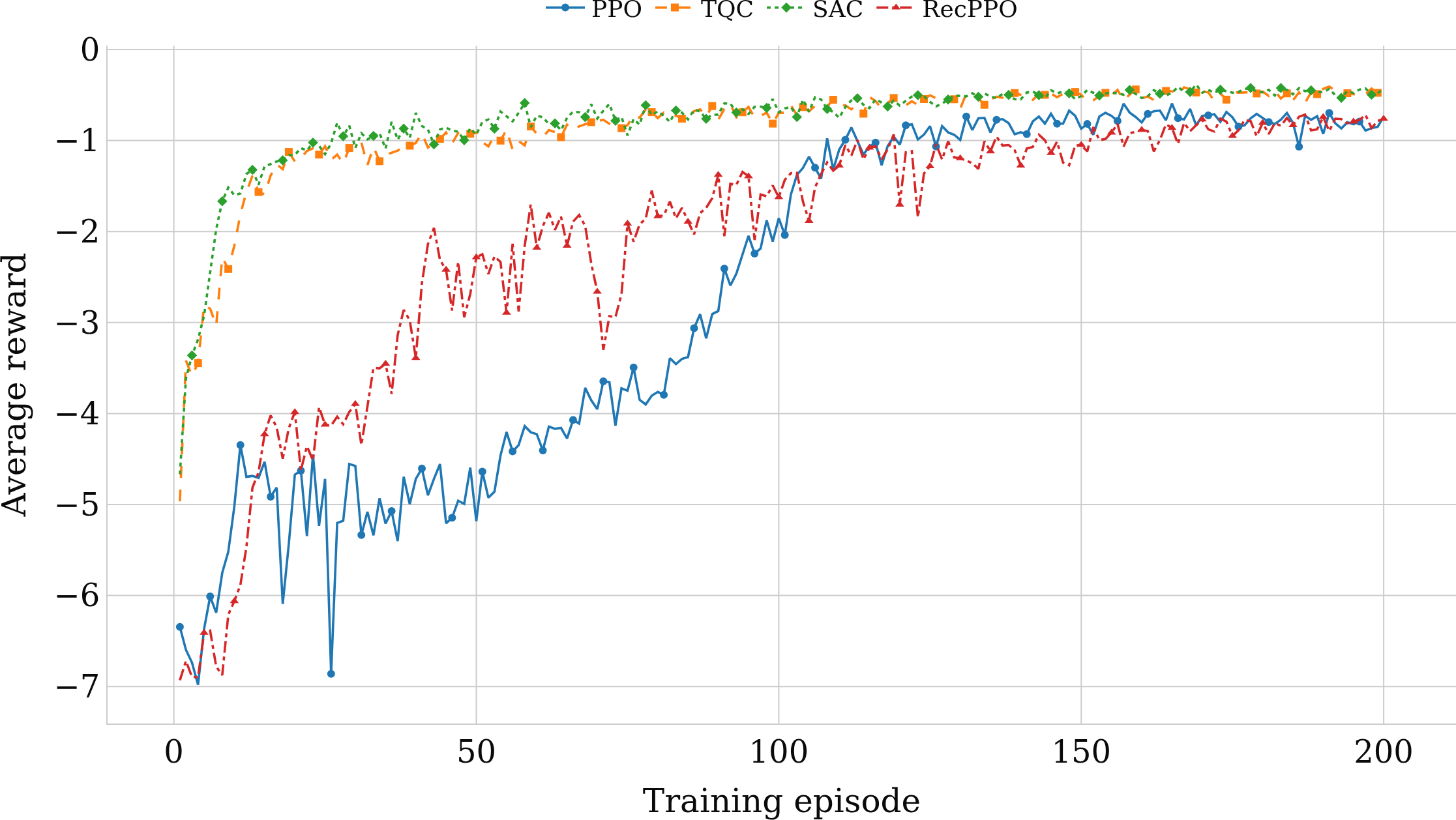}
    \caption{Reward evolution during training for different agents in the case study}
    \label{fig:case_study_reward}
\end{figure}

Figure~\ref{fig:case_study_reward} presents the reward evolution of the DRL agents during training. In this case, an energy weight of $\omega=0.75$ is considered. This weighting was selected based on several preliminary tests, which showed improved energy consumption without compromising thermal comfort. Furthermore, the number of training episodes was increased to 200 (compared with 100 in the pilot study) to ensure convergence of the DRL agents under the higher dimensionality of the control problem. It can be observed how algorithms converge to stable policies, indicating that the learning process remains robust despite the increased dimensionality of the control problem. Regarding convergence speed, the off-policy algorithms \texttt{TQC} and \texttt{SAC} again demonstrate higher sample efficiency, whereas \texttt{PPO} and \texttt{RecPPO} require a larger number of training episodes to achieve comparable performance, without eventually surpassing the performance of the off-policy approaches.

\begin{figure}[t]
    \centering
    \begin{subfigure}[b]{0.49\linewidth}
        \centering
        \includegraphics[width=\linewidth]{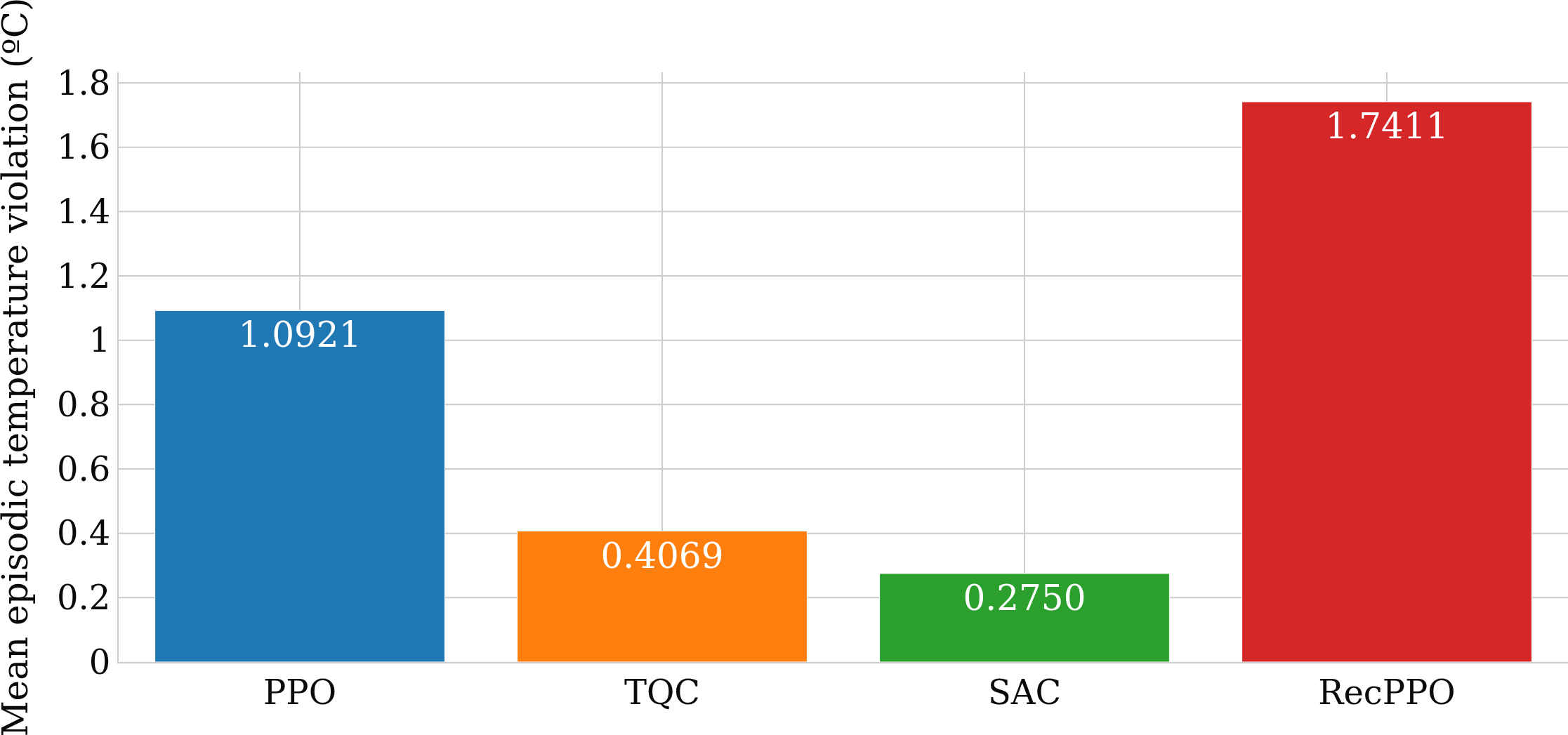}
        \caption{Mean episodic comfort violations}
        \label{fig:case_study_comfort_violations}
    \end{subfigure}
    \hfill
    \begin{subfigure}[b]{0.49\linewidth}
        \centering
        \includegraphics[width=\linewidth]{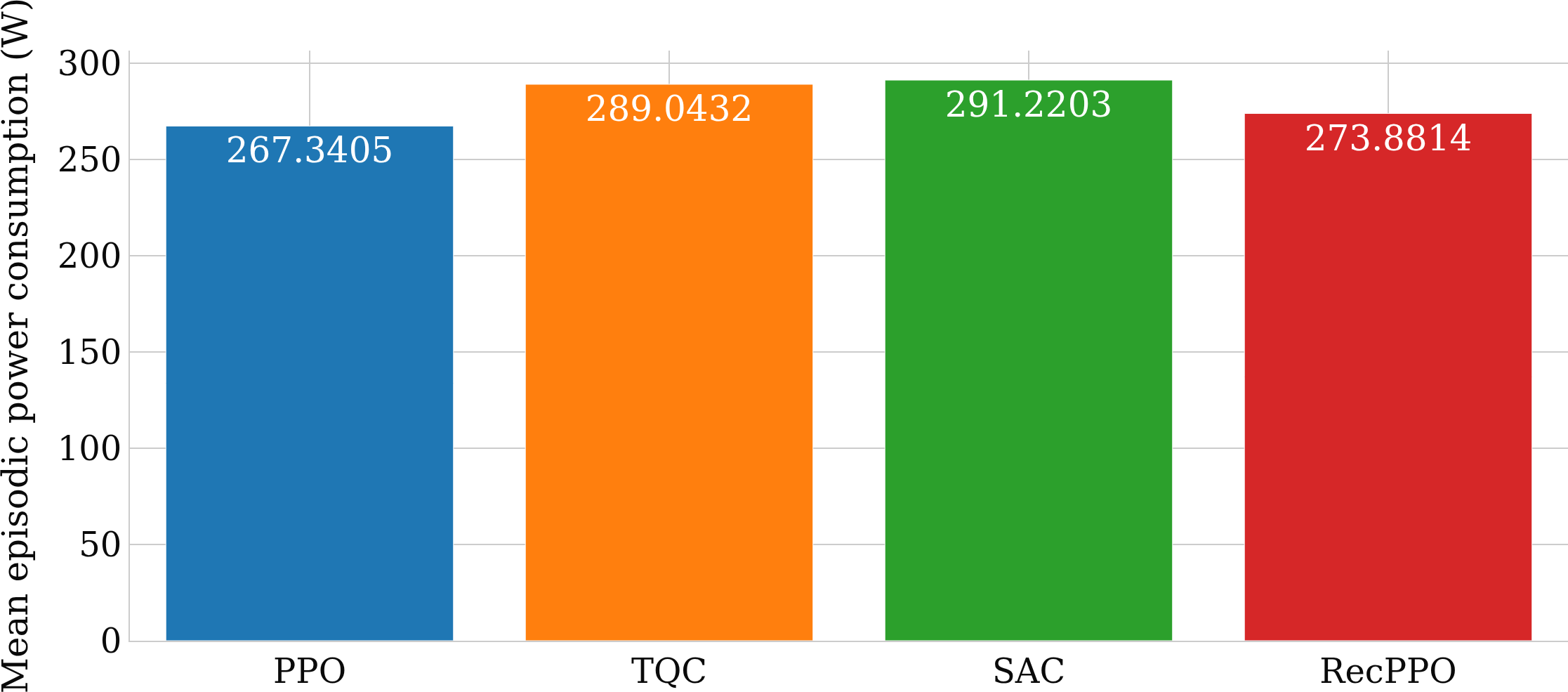}
        \caption{Mean episodic power consumption}
        \label{fig:case_study_power_demand}
    \end{subfigure}
    \caption{Comparison of the mean episodic performance of the DRL agents evaluated in the case study, in terms of comfort violations and power consumption (November 1st to March 31st)}
    \label{fig:case_study_mean_performance}
\end{figure}

The resulting control performance is summarized in Figure~\ref{fig:case_study_mean_performance}. Overall, all DRL agents are able to maintain indoor temperatures close to the desired comfort range. Once again, \texttt{TQC} and \texttt{SAC} agents achieve the best overall trade-off between thermal comfort and energy consumption, consistently outperforming the remaining algorithms. These results confirm the conclusions obtained during the pilot study, demonstrating that the selected control strategy successfully can scale to a substantially more complex building.

\begin{figure}
    \centering
    \includegraphics[width=\linewidth]{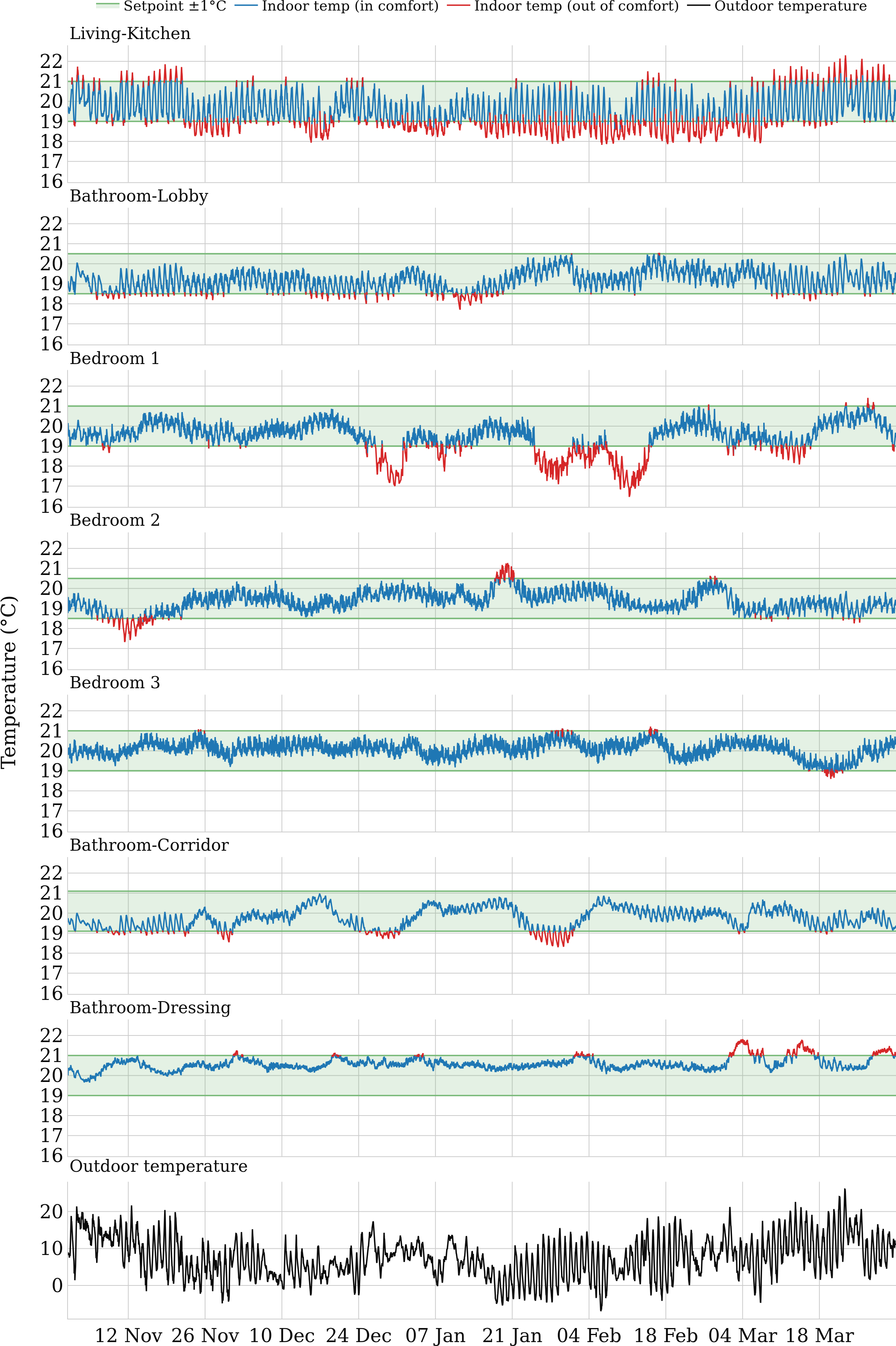}
    \caption{Indoor and outdoor temperatures evolution during a complete heating season in the simulated case-study building, evaluating the best-performing DRL agent (\texttt{TQC})}
    \label{fig:case_study_indoor_temp}
\end{figure}

Considering \texttt{TQC} as the best-performing DRL agent, Figure~\ref{fig:case_study_indoor_temp} illustrates the indoor temperature evolution achieved by this controller over a complete simulated heating season. The controller successfully maintains the indoor temperatures within the comfort band for the vast majority of the evaluation period while adapting the heating demand to the changing outdoor conditions. Most comfort violations occur in Bedroom~1, where the indoor temperature drops by up to 2ºC below the lower comfort limit during the coldest months. The remaining thermal zones exhibit more stable temperature profiles and only minor comfort violations. Having obtained these results in simulation, the \texttt{TQC} agent is subsequently deployed in the real-world environment for experimental validation.

\subsection{Simulation-to-reality transfer}

After selecting the best-performing controller in simulation, the trained \texttt{TQC} agent is deployed in the real building environment without any additional fine-tuning. The comfort ranges are defined according to the occupants' preferences, with independent values assigned to each thermal zone that can be adjusted over time. However, the occupants opted to maintain constant comfort ranges during most of the evaluation period, which spans from 24 March 2026 to 5 April 2026.

\begin{figure}
    \centering
    \includegraphics[width=\linewidth]{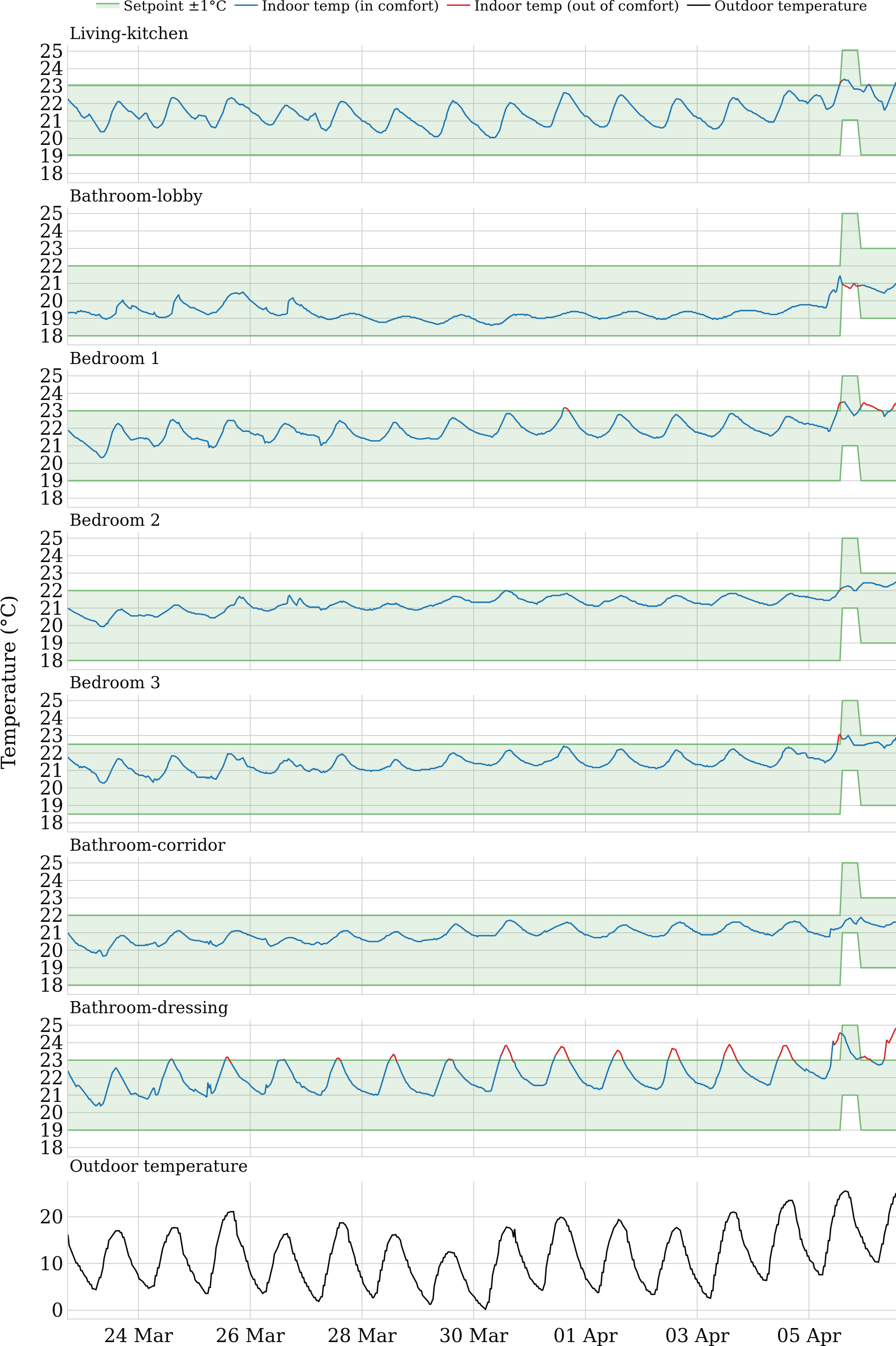}
    \caption{Indoor and outdoor temperatures evolution during real building deployment period (\texttt{TQC})}
    \label{fig:case_study_indoor_temp_real_home}
\end{figure}

Figure~\ref{fig:case_study_indoor_temp_real_home} shows the indoor temperature control achieved by the DRL agent. Overall, the agent provides adequate temperature regulation across all zones, with only minor comfort violations observed in the \textit{Bathroom-Dressing} zone. These violations are associated with brief peaks above the upper comfort limit, remaining below 1°C. 

\begin{figure}
    \centering
    \begin{subfigure}{\linewidth}
        \centering
        \includegraphics[width=\linewidth]{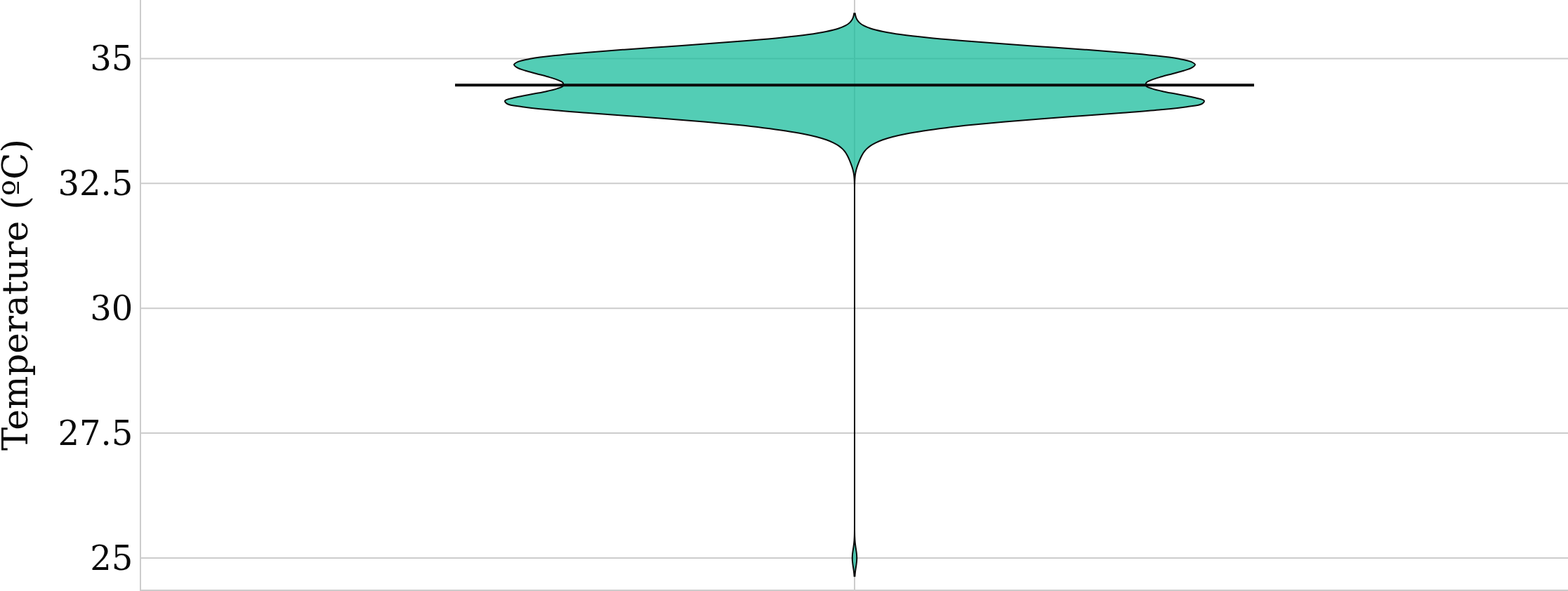}
        \caption{Supply temperature distribution}
        \label{fig:case_study_supply_temp_real_home}
    \end{subfigure}
    
    \vspace{0.5em}
    
    \begin{subfigure}{\linewidth}
        \centering
        \includegraphics[width=\linewidth]{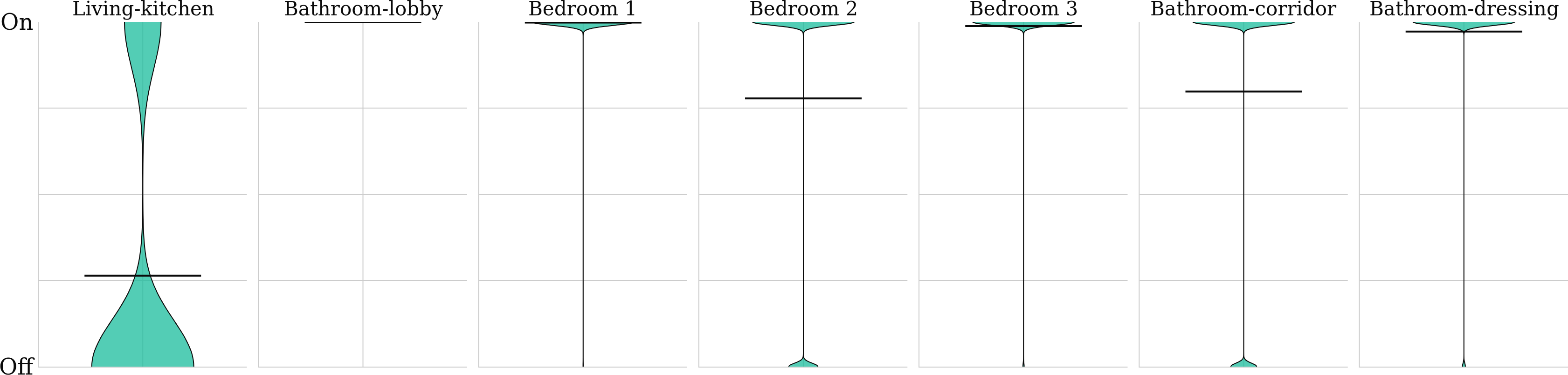}
        \caption{Valves opening distribution}
        \label{fig:case_study_valve_onoff_real_home}
    \end{subfigure}
    \caption{Action distributions during the real home deployment period (\texttt{TQC})}
    \label{fig:case_study_action_distributions_real_home}
\end{figure}

Regarding the agent's control policy, Figure~\ref{fig:case_study_supply_temp_real_home} shows that the supply temperature distribution remains relatively narrow and stable throughout the deployment period, with values centered around 34°C. This suggests that the controller avoids relying on abrupt variations in the HP operating point. On the other hand, Figure~\ref{fig:case_study_valve_onoff_real_home} shows that the living room exhibits the greatest variability in valve opening. This can be attributed to its larger size and greater exposure to solar radiation, both of which contribute to increased thermal variability and make the living room the room most strongly affected by solar gains. It can be observed that the valves remain open for most of the time in the majority of the circuits.

From the analysis of the temperature evolution in each zone, it can be observed that indoor temperatures frequently approach or exceed the upper comfort limit. By examining the agent's control actions, it is observed that this behaviour is not primarily caused by the controller itself. During these periods, the corresponding UFH valves remain fully closed while the indoor temperature continues to increase or decreases moderately slowly, indicating that the heating system is no longer supplying energy to these zones. This behaviour is mainly attributed to the effect of solar radiation.The deployment was conducted in late March and early April, a period characterized by relatively high solar irradiance in Madrid. The resulting passive solar gains have a significant impact on the thermal response of the dwelling, particularly in the most exposed rooms, leading to indoor temperatures exceeding the comfort band even in the absence of active heating. These observations highlight one of the main sources of discrepancy between the digital twin and the real building: the difficulty of accurately reproducing all solar gains and their interaction with the building envelope.

In conclusion, the main findings of the case study, covering both the training process in the digital twin and the subsequent real-world deployment, are as follows:

\begin{itemize}
    \item All evaluated DRL agents converged to stable control policies despite the increased dimensionality of the control problem. These results demonstrate that the proposed DRL-based control strategy successfully scales from the pilot building to a substantially larger and more complex residential building.
    
    \item \texttt{TQC} and \texttt{SAC} exhibited higher sample efficiency and achieved the best overall trade-off between thermal comfort and energy consumption, consistently outperforming \texttt{PPO} and \texttt{RecPPO}.
    
    \item The \texttt{TQC} agent, trained exclusively in the EnergyPlus digital twin, was successfully deployed in the real dwelling without additional fine-tuning. During the deployment period, the agent maintained indoor temperatures within the desired comfort ranges, with only limited comfort violations.
    
    \item During the real-world deployment, the heat pump supply temperature remained relatively stable, with values centered around 34°C, while the valve-opening distribution adapted to the thermal characteristics of each zone.
\end{itemize}


\section{Discussion, limitations, and practical considerations}
\label{sec:discussion}

\subsection{Discussion on the results}

Based on the obtained results, several key insights emerge regarding the practical implementation of DRL algorithms in real-world UFH systems. Notably, adopting a two-stage methodology ---starting with a simulation-based pilot study followed by training on a digital twin of the real building--- provided a flexible and versatile framework for experimental testing prior to physical deployment. This approach facilitated the evaluation of various reward functions, DRL agents, observation space configurations, and the required infrastructure for training and deployment. Furthermore, the pilot tests demonstrated the strong adaptability of DRL agents in maintaining thermal comfort even under degraded building envelope conditions, albeit at the cost of an adaptive increase in energy consumption to compensate for higher thermal losses.

Regarding algorithm performance, off-policy DRL algorithms (\texttt{TQC}, \texttt{SAC}) exhibited significantly higher sample efficiency and faster convergence rates compared to on-policy approaches (\texttt{PPO}, \texttt{RecPPO}). From a practical standpoint, this off-policy nature enables continuous offline re-training by progressively augmenting the experience replay buffer with real-world operational data collected during deployment. This strategy may mitigate the safety risks associated with online exploration in real systems while allowing the agent's policy to be iteratively refined without requiring training from scratch, thereby lowering deployment costs. Although still an active research avenue, leveraging such an off-policy approach holds strong potential for bridging the sim-to-real gap and enabling rapid policy adaptation \cite{blad2022data,ahmed2026dynamic,jang2026offline}.

When transitioning to the real-world deployment phase, the \texttt{TQC} agent ---trained exclusively on the EnergyPlus digital twin--- successfully maintained comfort temperatures across all thermal zones of the building. The learned control policy effectively stabilized the HP supply water temperature around 34~°C, preventing abrupt operational fluctuations. Combined with control actions executed at half-hour intervals, this favors continuous operation at less extreme supply temperatures over frequent start-stop cycles, thereby demonstrating potential for extending the equipment's service life. Simultaneously, the valve openings adapted dynamically according to the specific thermal demand and characteristics of each zone (e.g., showing greater variability in the living-kitchen area due to the higher impact of solar radiation).

\begin{figure}[htbp]
    \centering
    \includegraphics[width=\linewidth]{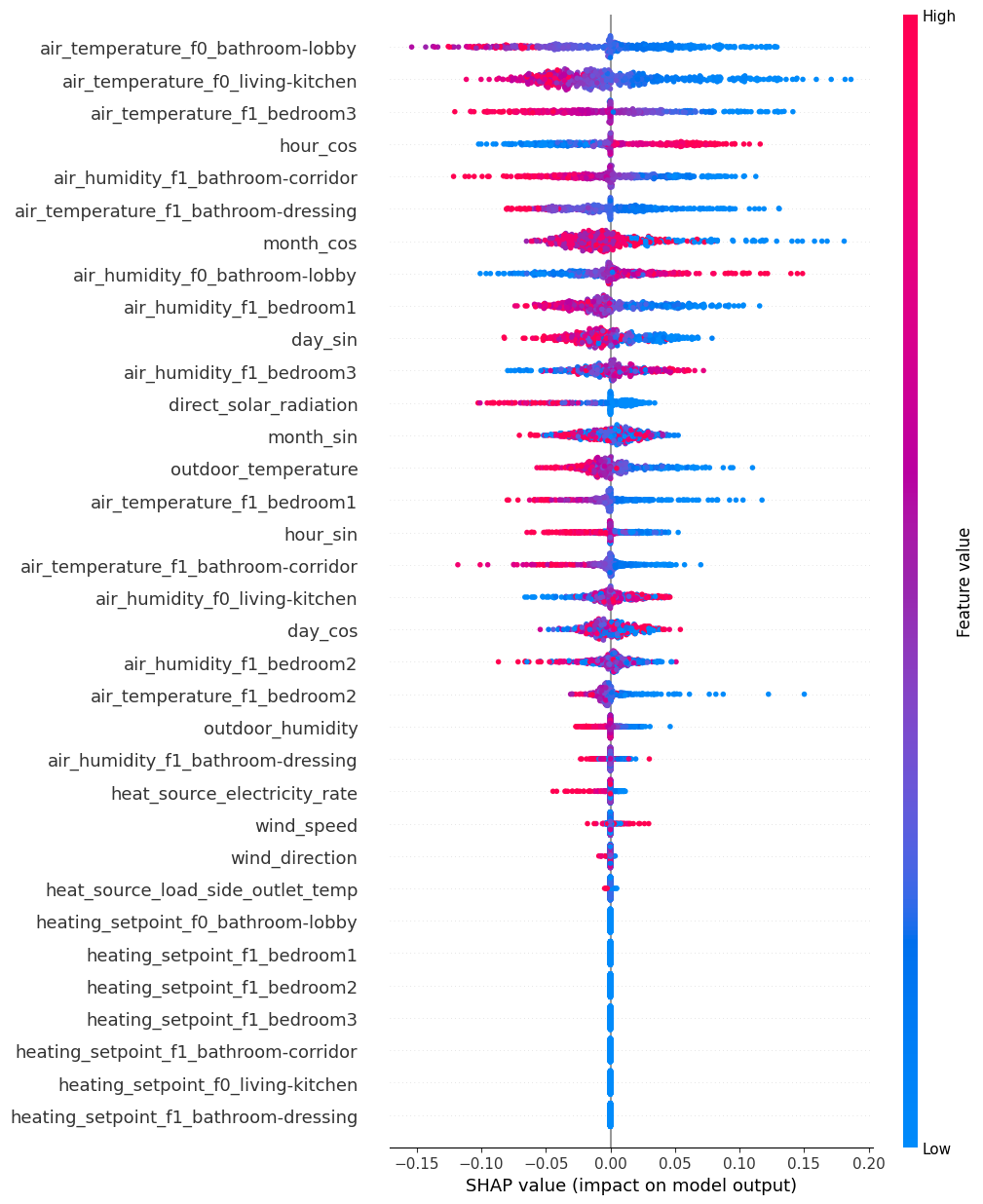}
    \caption{SHAP analysis of the trained \texttt{TQC} controller operating in the digital twin, illustrating the importance of the observed variables in the agent's outputs}
    \label{fig:shap_analysis}
\end{figure}

Despite these promising results, a primary drawback of DRL models remains their black-box nature, which hinders the interpretability of control actions and the influence of observed variables \cite{milani2024explainable,jimenez2023explaining}. To gain deeper insight into the agent's policy and enhance interpretability, Figure~\ref{fig:shap_analysis} illustrates a SHAP analysis \cite{lundberg2017unified} performed on the \texttt{TQC} policy operating within the digital twin. Aligned with the control performance shown in Figure~\ref{fig:case_study_indoor_temp}, the primary drivers behind the agent's decisions are indoor temperatures, relative humidity, and temporal features (hour, month, and day). Among exogenous variables, direct solar radiation exerts a stronger influence than outdoor temperature, reflecting the overheating conditions observed during operation. Conversely, factors such as wind speed and wind direction proved largely negligible to the agent. Heating setpoints also showed minimal impact, although this is attributable to the specific experimental setup, as they remained constant throughout the simulation.


\subsection{Limitations and practical considerations}

The deployment of the DRL controller in the real building revealed several practical limitations that were not fully captured during the simulation stages. These limitations arise from the interaction between the control algorithm, the UFH system, the building characteristics, and occupant behaviour, affecting both control reliability and the energy savings achievable in the real-world environment.


\paragraph{Communication infrastructure} One of the main practical challenges was the reliability of the actuation and communication infrastructure. The UFH valves were controlled using smart relays, enabling remote communication with the cloud-hosted controller. However, the wireless connection between the relays and the building's Wi-Fi network was a significant source of uncertainty. In conventional residential installations, the valve control system is typically installed inside a recessed metallic enclosure, which can substantially attenuate the Wi-Fi signal and cause intermittent loss of connectivity. In the proposed implementation, control actions are generated every 30 minutes, meaning that a disconnected relay can delay the valve state update until the next decision step. This delay creates a mismatch between the action selected by the DRL agent and the actual state of the physical system.


\paragraph{Action delays} An additional discrepancy between the simulation and the real implementation is related to the physical response time of the valves. In the simulation, control actions are assumed to take effect instantaneously, whereas the physical valves require approximately 1--2 minutes to complete their opening or closing operation after the corresponding electrical pulse is issued. This results in slightly slower transitions and, consequently, a slower thermal response. However, unlike the communication delay described above, this delay is considered negligible for the DRL controller, given the 30-minute interval between consecutive actions.


\paragraph{Thermal dynamics} The thermal characteristics of the building and weather conditions introduce an additional challenge for the control strategy. The ground floor comprises an open-plan living room and kitchen with a large glazed facade facing the garden, making solar radiation a major driver of its thermal dynamics. Its magnitude and temporal variability can cause significant changes in indoor temperature independently of the heating system operation, making the thermal response more difficult to predict and control. For the DRL controller, this represents a significant disturbance that can frequently lead to overheating. At the same time, accurately reproducing solar radiation and its interactions with multiple meteorological variables is challenging to reproduce in simulation, further increasing the complexity of training the DRL agent under realistic weather-variable distributions.


\paragraph{Occupant behaviour} Due to the thermal inertia of UFH systems, changes in heating setpoints do not immediately affect indoor temperature. However, occupants may not account for this inertia and can make successive manual adjustments when the expected temperature response is not immediate. These interventions may cause unnecessary heating operation, counteract the control policy selected by the DRL agent, and introduce additional variability that makes it difficult to distinguish the effects of the learned policy from those of human intervention. Although occupants were provided with recommendations on the operation of the system and the effects of its thermal inertia, consistent adherence proved challenging. This highlights a key difference between simulated and real-world RL environments: while the controller can generally assume exclusive control over the relevant variables in simulation, human actions introduce an external and potentially unpredictable component into the control loop.


\paragraph{Observation variables} The transition from the digital twin to the real building revealed several discrepancies in the distribution and availability of the observations used by the DRL agents. First, the deployment of the agents in the real building revealed that some of the sensor readings considered during simulation were not available in practice. Specifically, diffuse radiation and the heat source load-side inlet temperature could not be obtained from the available sensing infrastructure, as reflected by the differences between Tables~\ref{tab:variables_pilot} and \ref{tab:variables_case}. Consequently, the simulation phase had to be revisited and the agents retrained to account for the reduced observation space available during real-world deployment.

Furthermore, substantial discrepancies were identified between the external humidity and radiation measurements collected in the real building and the corresponding values provided by the EPW weather file used during simulation. Although the agents were still able to maintain coherent control behaviour, we decided to revisit the simulation and modify the distribution of the humidity and radiation observations to reduce the sim-to-real gap. In particular, the parameters of the Ornstein--Uhlenbeck process used to generate the corresponding disturbances were adjusted, including an increase in the overall mean of the simulated weather variables. This modification resulted in improved performance during deployment.

Nevertheless, significant discrepancies remained in the distribution of internal humidity. Further analysis showed that these differences were caused by an occupancy-related condition that had not been considered during the initial simulation setup, namely, the moisture released during occupant showers combined with situations in which bathroom doors were left open. To reproduce these abrupt humidity variations more accurately, a more substantial modification of the building model was required. Specifically, the model was extended to account for sudden moisture increases and to incorporate the propagation of humidity towards adjacent thermal zones. This additional modeling step reduced the discrepancy between the simulated and real humidity dynamics, thereby making the transition to real-world deployment less abrupt.


\paragraph{Heat pump cycling and system health} During the pilot study, different action frequencies were evaluated, including configurations with up to 12 agent actions per hour (i.e., one action every 5 minutes), which could potentially provide more fine-grained control of the system. However, increasing the action frequency did not result in a substantial improvement in agent performance. Moreover, frequent control actions represent a practical limitation for the operation of the HP, as they may lead to a higher number of compressor starts and, consequently, increased compressor cycling and system degradation. Therefore, the action frequency was ultimately limited to two actions per hour to avoid short cycling and preserve the operational health of the HP. This limitation is not considered detrimental to the control performance, since the system under study is a highly inertial thermal system. In particular, actions applied at shorter intervals may have only a limited effect on the floor temperature, as the thermal response of the system is considerably slower than the control timescale. 

Future work could explicitly incorporate HP health into the reward function by penalizing compressor starts or excessive cycling. This would extend the optimization objective beyond the conventional trade-off between thermal comfort and energy consumption, allowing the DRL agent to simultaneously account for the long-term health and degradation of the controlled system.


\paragraph{Seasonal testing and benchmarking} The available experimental period (late winter to early spring) coincided with mild outdoor temperatures and high solar gains. While this transition period presents complex predictive control challenges due to sharp diurnal temperature fluctuations, the agents could not be evaluated during the coldest months of the heating season. However, the primary objective of this real-world deployment was to demonstrate and validate the end-to-end functional integration of the DRL architecture in an occupied building. Further long-term tests are planned for upcoming heating seasons to assess the agent's robustness and thermal performance under extreme heating demand and peak cold conditions.

Simultaneously evaluating real-world deployment against baseline controllers (e.g., a standard heating curve or PI control) in an occupied residential building presents significant methodological challenges. Due to the high thermal inertia of the underfloor heating system, switching control strategies dynamically in a short time frame introduces thermal bias and potential discomfort for residents. Furthermore, direct real-world comparisons are constrained by changing weather and occupancy patterns. While the efficacy and savings of the proposed DRL framework compared to traditional baselines were thoroughly validated in the digital twin simulation phase, the field deployment prioritized testing the feasibility and resilience of the system under live operational constraints. Future work will extend this deployment to enable systematic, multi-season benchmarking against baseline strategies under strictly comparable operating conditions.


\section{Conclusions and future work}
\label{sec:conclusions}

This paper demonstrates the feasibility of deploying a DRL-based controller for the control of UFH systems coupled with HPs in residential buildings. First, a series of experiments were conducted using a generic residential building model to evaluate and compare different algorithms, control strategies, and training configurations. Based on the results obtained, a real-world control case was subsequently deployed and evaluated in an occupied residential building. Several DRL agents were trained using a digital twin of the building, and the best-performing agent was then deployed in the real building, enabling its performance to be evaluated under real operating conditions.

The results in terms of thermal comfort and energy consumption indicate that off-policy algorithms, particularly \texttt{TQC} and \texttt{SAC}, achieved the most favorable performance among the evaluated approaches. The \texttt{TQC} agent, trained exclusively using an EnergyPlus-based digital twin, was successfully transferred to the real building without prior fine-tuning. During the evaluation period, the agent maintained indoor temperatures within the desired comfort ranges, demonstrating the potential of simulation-to-real transfer for DRL-based HVAC control.

Additional experiments were also conducted to assess the robustness and interpretability of the proposed controllers. Robustness tests were performed to evaluate the ability of the agents to adapt to degraded building envelopes and changes in building performance. Furthermore, to provide greater insight into the decisions made by the resulting black-box policies, the influence of relevant variables on the agents' actions was analyzed. These variables included indoor temperature, solar radiation, outdoor temperature, and time-related features.

Overall, this work represents an experimental contribution to the application of DRL to cyber-physical systems, particularly in the HVAC domain, where a large proportion of the reported results are still evaluated exclusively in simulation and only a limited number of studies are validated in real buildings. Nevertheless, further research is required to assess the scalability and generalization capabilities of these controllers across different building configurations, weather conditions, and HVAC systems. Future work will therefore focus on extending the proposed approach to larger and more diverse building sets and to alternative HVAC systems. Comparative analyses with other control approaches will also be conducted during subsequent heating seasons to further assess the performance and robustness of the DRL controllers. In addition, the proposed control framework will be extended to the cooling season, including its integration with fan-coil units.

In conclusion, the combination of digital-twin-based training, real-world deployment, robustness assessment, and analysis of decision-relevant variables provides a comprehensive framework for evaluating DRL controllers beyond purely simulation-based studies. This work represents a step forward in bridging the gap between simulation and real-world deployment, supporting the implementation of efficient control strategies for building energy systems.


\section*{Data availability}

The datasets used in this study are available in the following repository: \url{http://rl-datasets.gf-digitalcontrols.com/paper.zip}.





\section*{Declaration of generative AI in the manuscript preparation process}

During the preparation of this work, the authors used Artificial intelligence (AI)-based tools in order to assist with the linguistic revision and translation of the manuscript. After using these tools, the authors reviewed and edited the content as needed. All scientific content, interpretations, analyses, and conclusions were developed and critically reviewed by the authors, who remain fully responsible for the final content of the manuscript.


\appendix

\section{Observed variables}
\label{app:variables}

Tables~\ref{tab:variables_pilot} and \ref{tab:variables_case} describe the variables that constitute the observation space of the agents for the pilot and case studies, respectively. It should be noted that the number of observed variables in the case study is lower than in the pilot study due to the limited availability of sensors in the real building. Using the wrappers provided by Sinergym, all variables are normalized, while temporal variables (day, month, and hour) are encoded using sine and cosine transformations.

\begin{table}[ht]
    \centering
    \footnotesize
    \begin{tabular}{c l l c}
        \toprule
        {} & \textbf{Variable} & \textbf{Description} & \textbf{Unit} \\
        \midrule
        1 & \texttt{month\_cos} & Cosine encoding of the month & - \\
        2 & \texttt{month\_sin} & Sine encoding of the month & - \\
        3 & \texttt{day\_cos} & Cosine encoding of the day & - \\
        4 & \texttt{day\_sin} & Sine encoding of the day & - \\
        5 & \texttt{hour\_cos} & Cosine encoding of the hour & - \\
        6 & \texttt{hour\_sin} &  Sine encoding of the hour & - \\
        \midrule
        7 & \texttt{outdoor\_temperature} & Outdoor temperature & \textcelsius \\
        8 & \texttt{outdoor\_humidity} & Outdoor relative humidity & \% \\
        9 & \texttt{wind\_speed} & Wind speed & m/s \\
        10 & \texttt{wind\_direction} & Wind direction & $^\circ$ \\
        11 & \texttt{diffuse\_solar\_radiation} & Diffuse solar radiation & W/m$^2$ \\
        12 & \texttt{direct\_solar\_radiation} & Direct solar radiation & W/m$^2$ \\
        \midrule
        13 & \texttt{air\_temperature\_X} & Air temperature in zone \texttt{X} & \textcelsius \\
        14 & \texttt{air\_humidity\_X} & Relative humidity in zone \texttt{X} & \% \\
        15 & \texttt{heating\_setpoint\_X} & Heating setpoint in zone \texttt{X} & \textcelsius \\
        \midrule
        16 & \texttt{heat\_source\_load\_side\_outlet\_temp} & Heat source load side outlet temperature & \textcelsius \\
        17 & \texttt{heat\_source\_load\_side\_inlet\_temp} & Heat source load side inlet temperature & \textcelsius \\
        18 & \texttt{heat\_source\_electricity\_rate} & Heat source electricity consumption & W \\
        \bottomrule
    \end{tabular}
    \caption{Observed variables in the pilot study}
    \label{tab:variables_pilot}
\end{table}

\begin{table}[ht]
    \centering
    \footnotesize
    \begin{tabular}{c l l c}
        \toprule
        {} & \textbf{Variable} & \textbf{Description} & \textbf{Unit} \\
        \midrule
        1 & \texttt{month\_cos} & Cosine encoding of the month & - \\
        2 & \texttt{month\_sin} & Sine encoding of the month & - \\
        3 & \texttt{day\_cos} & Cosine encoding of the day & - \\
        4 & \texttt{day\_sin} & Sine encoding of the day & - \\
        5 & \texttt{hour\_cos} & Cosine encoding of the hour & - \\
        6 & \texttt{hour\_sin} &  Sine encoding of the hour & - \\
        \midrule
        7 & \texttt{outdoor\_temperature} & Outdoor temperature & \textcelsius \\
        8 & \texttt{outdoor\_humidity} & Outdoor relative humidity & \% \\
        9 & \texttt{wind\_speed} & Wind speed & m/s \\
        10 & \texttt{wind\_direction} & Wind direction & $^\circ$ \\
        11 & \texttt{direct\_solar\_radiation} & Direct solar radiation & W/m$^2$ \\
        \midrule
        12 & \texttt{air\_temperature\_X} & Air temperature in zone \texttt{X} & \textcelsius \\
        13 & \texttt{air\_humidity\_X} & Relative humidity in zone \texttt{X} & \% \\
        14 & \texttt{heating\_setpoint\_X} & Heating setpoint in zone \texttt{X} & \textcelsius \\
        \midrule
        15 & \texttt{heat\_source\_load\_side\_outlet\_temp} & Heat source load side outlet temperature & \textcelsius \\
        16 & \texttt{heat\_source\_electricity\_rate} & Heat source electricity consumption & W \\
        \bottomrule
    \end{tabular}
    \caption{Observed variables in the case study}
    \label{tab:variables_case}
\end{table}

\FloatBarrier


\section{Building degradation configurations}
\label{app:degradations}

To evaluate the impact of building aging on the proposed methodology, several degradation scenarios were defined. These scenarios emulate deterioration of the building envelope by modifying material properties, window thermal performance, and infiltration characteristics. 

Table~\ref{tab:material_degradation_summary} lists those materials for which a degraded version was defined, with a summary of the modifications applied to their thermophysical properties to emulate aging and deterioration of the building envelope. A complete definition of these materials can be consulted in \cite{energyplusmaterials}. Tables~\ref{tab:window_degradation} and \ref{tab:infiltration_degradation} summarize the building degradation scenarios considered for both windows and wall infiltrations. These degradation scenarios were applied to the simulated building both individually and in combination. The combined degradation scenarios are presented in Table~\ref{tab:combined_degradations}.





\begin{table}[h]
    \centering
    \caption{Building envelope material degradations considered}
    \label{tab:material_degradation_summary}

    \begin{subtable}[h]{0.48\textwidth}
        \centering
        \footnotesize
        \caption{Degraded building materials}
        \label{tab:degraded_materials}
        \renewcommand{\arraystretch}{1.2}
        \begin{tabular}{l}
            \toprule
            \textbf{Material} \\
            \midrule
            Gypsum Board (8 mm) \\
            Exterior Brick Masonry (100 mm) \\
            Dense Cast Concrete (100 mm) \\
            Lightweight Cast Concrete (100 mm) \\
            Standard Cast Concrete (100 mm) \\
            Ceramic/Porcelain Tile (9.5 mm) \\
            Medium-Density Concrete Block (100 mm) \\
            Expanded Polystyrene Insulation (80 mm) \\
            Gypsum Plasterboard (25 mm) \\
            Gypsum Plaster (13 mm) \\
            Soft Limestone (45 mm) \\
            Glass Wool Roof Insulation \\
            Plasterboard (13 mm) \\
            Heavyweight Plywood (10 mm) \\
            Urea-Formaldehyde Foam Insulation (80 mm) \\
            Extruded Polystyrene (XPS) Wall Insulation \\
            \bottomrule
        \end{tabular}
    \end{subtable}
    \hfill
    \begin{subtable}[h]{0.48\textwidth}
        \centering
        \footnotesize
        \caption{General degradations applied to the building envelope}
        \label{tab:material_degradation}
        \renewcommand{\arraystretch}{1.2}
        \begin{tabular}{lc}
            \toprule
            \textbf{Property} & \textbf{Applied modification} \\
            \midrule
            Thermal conductivity & Increased by 10--20\% \\
            Thickness & Reduced by 5--10\% \\
            Thermal absorptance & Reduced from 0.90 to 0.85 (5\%) \\
            Solar absorptance & Reduced by 5\% \\
            Visible absorptance & Reduced by 5\% \\
            \bottomrule
        \end{tabular}
    \end{subtable}

\end{table}

\begin{table}
    \centering
    \footnotesize
    \caption{Window degradation configurations}
    \label{tab:window_degradation}
    \renewcommand{\arraystretch}{1.2}
    \begin{tabular}{lcc}
        \toprule
        \textbf{Name} & \textbf{Solar Heat Gain Coefficient} & \textbf{U-factor} \\
        \midrule
        \texttt{Default}   & 0.40 & 1.40 \\
        \texttt{Window\_1}  & 0.85 & 5.80 \\
        \texttt{Window\_2}  & 0.30 & 1.80 \\
        \texttt{Window\_3}  & 0.30 & 2.50 \\
        \texttt{Window\_4}  & 0.50 & 2.50 \\
        \bottomrule
    \end{tabular}
\end{table}

\begin{table}
    \centering
    \footnotesize
    \caption{Wall infiltration values}
    \label{tab:infiltration_degradation}
    \renewcommand{\arraystretch}{1.2}
    \begin{tabular}{lcc}
        \toprule
        \textbf{Name} & \textbf{Air Changes per Hour} & \textbf{Temperature Term Coefficient} \\
        \midrule
        \texttt{Default}         & 0.10 & 0.000 \\
        \texttt{Infiltration\_1}  & 0.30 & 0.000 \\
        \texttt{Infiltration\_2}  & 0.30 & 0.050 \\
        \texttt{Infiltration\_3}  & 0.20 & 0.010 \\
        \texttt{Infiltration\_4}  & 0.10 & 0.050 \\
        \texttt{Infiltration\_5}  & 0.20 & 0.025 \\
        \bottomrule
    \end{tabular}
\end{table}

\begin{table}
    \centering
    \footnotesize
    \caption{Combined degradation scenarios, considering: all material degradation, air infiltrations and degraded window insulation}
    \label{tab:combined_degradations}
    \renewcommand{\arraystretch}{1.2}
        \begin{tabular}{ll}
        \toprule
        \textbf{Name} & \textbf{Combination} \\
        \midrule
        \texttt{All\_1} & \texttt{Material} + \texttt{Infiltration\_5} + \texttt{Window\_2} \\
        \texttt{All\_2} & \texttt{Material} + \texttt{Infiltration\_3} + \texttt{Window\_2} \\
        \bottomrule
    \end{tabular}
\end{table}

\FloatBarrier


\bibliographystyle{elsarticle-num} 
\bibliography{bibliography}

\end{document}